\documentclass[10pt,journal,compsoc]{IEEEtran}
\usepackage{hyperref}
\usepackage{color}
\usepackage[fleqn]{amsmath}
\usepackage{graphicx}
\usepackage{epigraph}
\usepackage{pifont}
\usepackage{amssymb}
\usepackage{subfig}
\usepackage{rotating}
\usepackage{adjustbox}
\usepackage{booktabs}
\usepackage{multirow}
\usepackage{tabularx}

\newcommand{\cmark}{\ding{51}}%
\newcommand{\xmark}{\ding{55}}%

\usepackage{array}
\newcolumntype{L}[1]{>{\raggedright\let\newline\\\arraybackslash\hspace{0pt}}m{#1}}
\newcolumntype{C}[1]{>{\centering\let\newline\\\arraybackslash\hspace{0pt}}m{#1}}
\newcolumntype{R}[1]{>{\raggedleft\let\newline\\\arraybackslash\hspace{0pt}}m{#1}}

\ifCLASSINFOpdf
\else
\fi

\newcommand{\startw}{{$\star\star$}}
\newcommand{\starth}{{$\star\star$$\star$}}

\begin{document}


\title{Indoor Scene Understanding in 2.5/3D for Autonomous Agents: A Survey}

\author{Muzammal Naseer$^{\dagger}$, Salman H. Khan$^{\star}$$^{\ddagger}$, Fatih Porikli$^{\dagger}$   \\
 $^{\dagger}$Australian National University, $^{\star}$Data61-CSIRO, $^{\ddagger}$Inception Institute of AI\\
{\tt\small muzammal.naseer@anu.edu.au}
}




\IEEEcompsoctitleabstractindextext{
\begin{abstract} 
With the availability of low-cost and compact 2.5/3D visual sensing devices, computer vision community is experiencing a growing interest in visual scene understanding of indoor environments. This survey paper provides a comprehensive background to this research topic. We begin with a historical perspective, followed by popular 3D data representations and a comparative analysis of available datasets. Before delving into the application specific details, this survey provides a succinct introduction to the core technologies that are the underlying methods extensively used in the literature. Afterwards, we review the developed techniques according to a taxonomy based on the scene understanding tasks. This covers holistic indoor scene understanding as well as subtasks such as scene classification, object detection, pose estimation, semantic segmentation, 3D reconstruction, saliency detection, physics-based reasoning and affordance prediction. Later on, we summarize the performance metrics used for evaluation in different tasks and a quantitative comparison among the recent state-of-the-art techniques. We conclude this review with the current challenges and an outlook towards the open research problems requiring further investigation. 
\end{abstract}

}
\IEEEdisplaynotcompsoctitleabstractindextext
\maketitle

     
\section{Introduction}

\setlength{\epigraphwidth}{0.4\textwidth}
\epigraph{It's not what you look at that matters, it's what you see.}{\textit{H.D. Thoreau (1817-62)}}

\IEEEPARstart{A}{n} image is simply a grid of numbers to a machine. In order to develop a comprehensive understanding of visual content, it is necessary to uncover the underlying geometric and semantic clues. As an example, given an RGB-D (2.5D) indoor scene, a vision-based AI agent should be able to understand the complete 3D spatial layout, functional attributes and semantic labels of the scene and its constituent objects. Furthermore, it is also required to comprehend both the apparent and hidden relationships present between scene elements. These capabilities are fundamental to the way humans perceive and interpret images, thus imparting these astounding abilities in machines has been a long-standing goal in computer vision discipline.
We can formally define visual scene understanding problem in machine vision as follows:

\begin{quotation}
\noindent
\textbf{Scene Understanding: } \textit{``To analyze a scene by considering the geometric and semantic context of its contents and the intrinsic relationships between them.''}
\end{quotation}

Visual scene understanding can be broadly divided into two categories based on the input media: static (for an image) and dynamic (for a video) understanding. This survey specifically attends to static scene understanding of 2.5/3D visual data  for indoor scenes. We focus on the 3D media since the 3D scene understanding capabilities are central to the development of general-purpose AI agents that can be deployed for emerging application areas as diverse as autonomous vehicles \cite{geiger2012we}, domestic robotics \cite{breuer2012johnny}, health-care systems \cite{teistler2003virtual}, education \cite{billinghurst2012augmented}, environment preservation \cite{widodo2000vehicle} and infotainment \cite{Qual3D}. According to an estimate from WHO, there are $253$ million people suffering from vision impairment \cite{WHO}. 3D scene understanding can help them safely navigate by detecting obstacles and analyzing the terrain \cite{rodriguez2012obstacle}. Domestic robots with cognitive abilities can be used to take care of elderly people, whose number is expected to reach 1.5 billion by the year 2050.

\begin{figure*}[ht]
  \centering
 \subfloat[Scene Classification]{\label{scene_1}\includegraphics[width=0.23\linewidth, height=3.1cm]{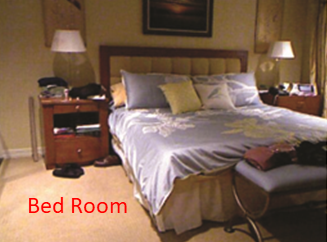}}
 \hspace{0.1cm}
 \subfloat[Semantic Segmentation]{\label{scene_2}\includegraphics[width=0.23\linewidth, height=3.1cm]{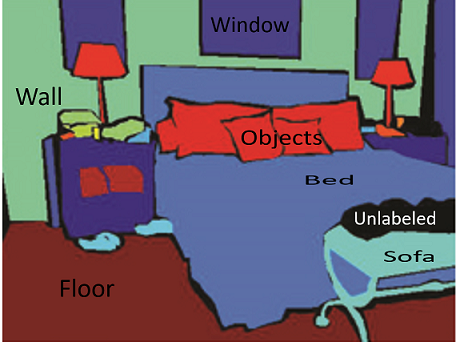}}
 \hspace{0.1cm}
 \subfloat[Object Detection]{\label{scene_3}\includegraphics[width=0.23\linewidth, height=3.1cm]{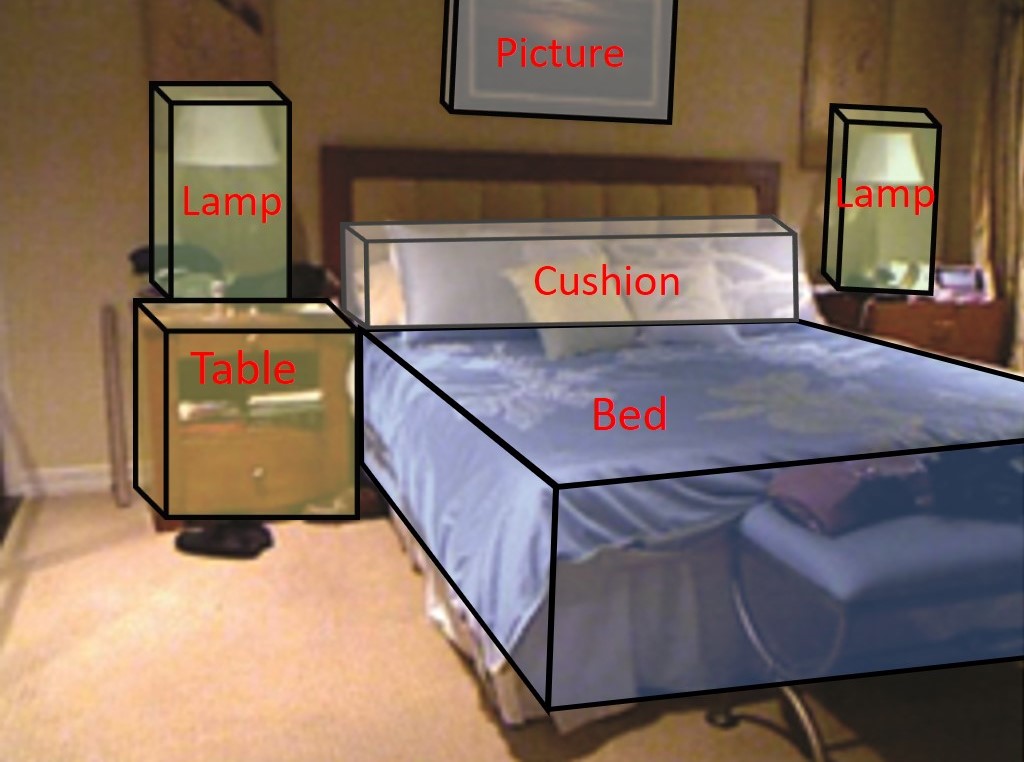}}
  \hspace{0.1cm}
 \subfloat[Pose Estimation]{\label{scene_4}\includegraphics[width=0.23\linewidth, height=3.1cm]{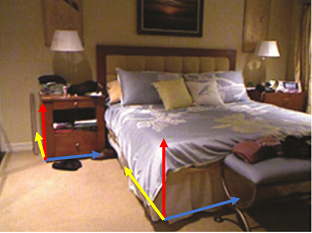}}
 \\
  \subfloat[Physics based reasoning]{\label{scene_5}\includegraphics[width=0.23\linewidth, height=3.1cm]{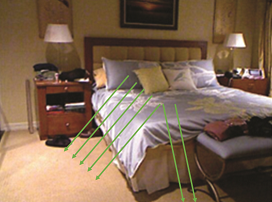}}
 \hspace{0.1cm}
 \subfloat[Saliency Prediction]{\label{scene_6}\includegraphics[width=0.23\linewidth, height=3.1cm]{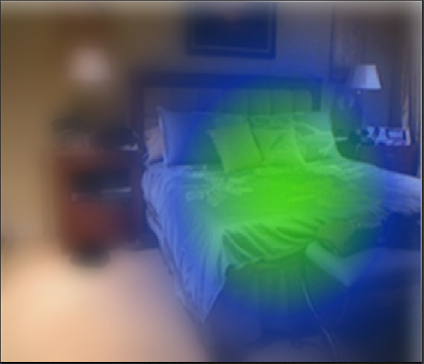}} 
  \hspace{0.1cm}
 \subfloat[Affordance Prediction]{\label{scene_7}\includegraphics[width=0.23\linewidth, height=3.1cm]{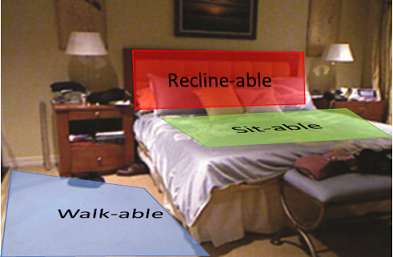}}
  \hspace{0.1cm}
 \subfloat[3D Reconstruction \cite{song2016semantic}]{\label{scene_8}\includegraphics[width=0.23\linewidth, height=3.1cm]{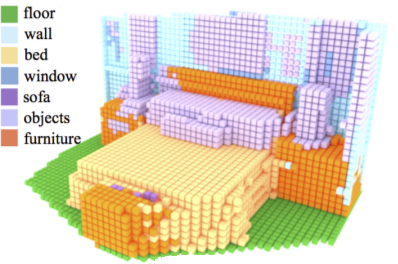}}
 
\caption{Given a RGB-D image, visual scene understanding can involve the image and pixel level semantic labeling (a-b), 3D object detection and pose estimation (c-d), inferring physical relationships (e), identifying salient regions (f), predicting affordances (g), full 3D reconstruction (h) and holistic reasoning about multiple such tasks (sample image from the NYU-Depth dataset \cite{NYUDv2}).}
 \label{scene}
\end{figure*}

As much as being highly significant, 3D scene understanding is also remarkably challenging due to the complex interactions between objects, heavy occlusions, cluttered indoor environments, major appearance, viewpoint and scale changes across different scenes and the inherent ambiguity in the limited information provided by a static scene. Recent developments in large-scale data-driven models, fueled by the availability of big annotated datasets have sparked a renewed interest in addressing these challenges. This survey aims to provide an inclusive background to this field, with a review of the competing methods developed recently. Our intention is not only to explore the existing wealth of knowledge but also to identify the key areas lacking substantial interest from the community and the potential future directions crucial for the development of practical AI-based systems. To this end, we cover both the specific problem domains under the umbrella of scene understanding as well as the underlying computational tools that have been used to develop state-of-the-art solutions to various scene analysis problems (Fig.~\ref{scene}). To the best of our knowledge, this is the first review that broadly summarizes the progress and promising new directions in 2.5/3D indoor scene understanding. We believe this contribution will serve as a helpful reference to the community.

\section{A Brief History of 3D Scene Analysis }

There exists a fundamental difference in the way a machine and a human would perceive the visual content. An image or a video is, in essence, a tensor with numeric values representing color (e.g., $r, g$ and $b$ channels) or location (e.g., $x, y$ and $z$ coordinates) information. An obvious way of processing such information is to compute local features representing color and texture characteristics. To this end, a number of local feature descriptors have been designed over the years to faithfully encode visual information. Some of these include e.g., SIFT \cite{lowe1999object}, HOG \cite{dalal2005histograms}, SURF \cite{bay2006surf}, Region Covariance \cite{tuzel2006} and LBP \cite{ahonen2006face} to name a few. The human visual system not only perceives the local visual details but also cognitively reasons about semantics and geometry in a scene, and can understand complex relationships between objects. Efforts have been made to replicate these remarkable visual capabilities in machine vision for advanced applications such as context-aware personal digital assistants, health-care and domestic robotic systems, content-driven retrieval and assistive devices for visually impaired.

Initial work on scene understanding was motivated by the human cognitive psychology and neuroscience.  In this regard, several notable ideas were put forward to explain the working of the human visual system. In 1867, Helmholtz \cite{charlton1968helmholtz} explained his concept of `unconscious conclusion', which attributes the involuntary visual perception to our longstanding previous interactions with the 3D surroundings. In 1920s, Gestalt theory argued that the holistic interpretation of a scene developed by humans is due to eight main factors, the prominent ones being proximity, closure, and common motion \cite{koffka2013principles}.  Barrow and Tenenbaum \cite{barrow1978computer} introduced the idea of `intrinsic images', which are layers of visual information a human can easily extract from a given scene. These include illumination, reflectance, depth, and orientation. Around half a century ago, Marr proposed his three-level vision theory, which transitions from a 2D primal sketch of a scene (consisting of edges and regions), first to a 2.5D sketch (consisting of texture and orientations) and finally to a 3D model which encodes complete shape of a scene \cite{marrcomputational}. 

\begin{figure*}[!htp]
 \centering
 \subfloat[CAD Model]{\label{figur:11}\includegraphics[width=0.15\linewidth]{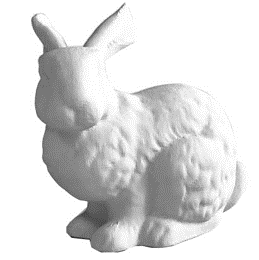}}
 \subfloat[ Point Cloud]{\label{figur:12}\includegraphics[width=0.15\linewidth]{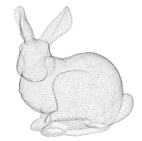}}
 \subfloat[Mesh]{\label{figur:13}\includegraphics[width=0.12\linewidth]{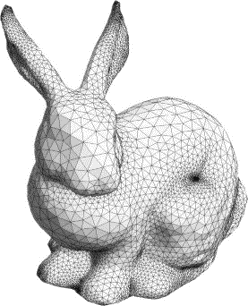}}
 \subfloat[Voxelized]{\label{figur:14}\includegraphics[width=0.145\linewidth]{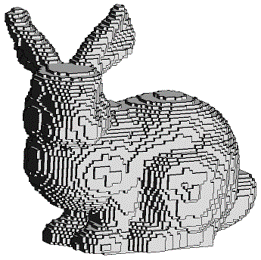}}
  \subfloat[Octree]{\label{figur:15}\includegraphics[width=0.14\linewidth]{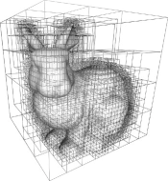}}
 \subfloat[TSDF]{\label{figur:16}\includegraphics[width=0.3\linewidth]{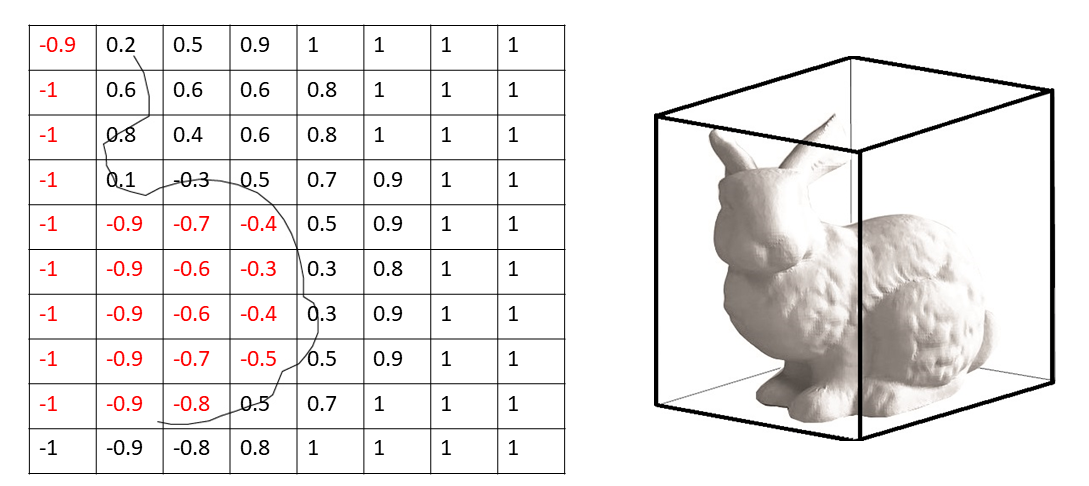}}
 \caption{Visualization of different types of 3D data representations for Stanford bunny. }
\label{fig:data_vis}
\end{figure*}

Representation is a key element of understanding the 3D world around us. In the early days of computer vision, researchers favored parts-based representations for object description and scene understanding. One of the initial efforts in this regard was made by L.G. Roberts \cite{roberts1963machine}, who presented an approach to denote objects using a set of 3D polyhedral shapes. Afterwards, a set of object parts was identified by A. Guzman \cite{guzman1968decomposition} as the primitives for representing generic 2D shapes in line-drawings. Another seminal idea was put forward by T. Binford, who demonstrated that several curved objects could be represented using generalized cylinders \cite{binford1971visual}. Based on the generalized cylinders, a pioneering contribution was made by I. Biederman, who introduced a set of basic primitives (termed as `geons' meaning geometrical ions) and linked it with the object recognition in human cognitive system \cite{biederman1985human}. Recently, data-driven feature representations learned using deep neural networks have been shown to perform superior for describing visual data \cite{song2016deep,zhang2016deepcontext,girshick2015fast,simonyan2014very}.


While the initial systems developed for scene analysis bear notable ideas and insights, they lack generalizability to new scenes. This was mainly caused due to handcrafted rules and brittle logic-based pipelines. Recent advances in automated scene analysis seek to resolve these issues by devising more flexible, learning based approaches that offer rich expressiveness, efficient training, and inference in the designed models. We will systematically review the recent approaches and core tools in Sec.~\ref{sec:core} and \ref{sec:sub-problems}. However, before that, we provide an overview of the underlying data representations and datasets for RGB-D and 3D data in the next two sections.  

\section{Data Representations}

In the following, we highlight the popular 2.5D and 3D data representations used to represent and analyze scenes. An illustration of different representations is provided in Fig.~\ref{fig:data_vis}, while a comparative analysis is reported in Table~\ref{tab:data_rep}.

\noindent
$\bullet$ \textbf{Point Cloud:}
A `point cloud' is a collection of data points in 3D space. The combination of these points can be used to describe the geometry of the individual object or the complete scene. Every point in the point cloud is defined by $x$, $y$ and $z$ coordinates, which denote the physical location of the point in 3D. Range scanners (typically based on laser, e.g., LiDAR) are also used to capture 3D point clouds of objects or scenes. 

\noindent
$\bullet$ \textbf{Voxel Representation:}
A voxel (volumetric element) is the 3D counterpart of a pixel (picture element) in a 2D image. Voxelization is a process of converting a continuous geometric object into a set of discrete voxels that best approximate the object. A voxel can be considered as a cubic volume representing a unit sample on a uniformly spaced 3D grid. Usually, a voxel value is mapped to either 0 or 1, where 0 indicates an empty voxel while 1 indicates the presence of range points inside the voxel.


\begin{table}
\setlength\tabcolsep{1.2pt}
\centering
\begin{tabular}{c c c c c}
\toprule
Representation & Data  & Memory  & Shape  & Computation \\
 & Dimension & Efficiency &  Details & Efficiency\\
\midrule
Point cloud & 3D & \startw  & \starth & \starth \\
Voxel & 3D & \starth & \startw & \starth \\
Mesh & 3D & $\star$ & \starth & \startw \\
Depth & 2.5D & \starth & $\star$ & \starth \\
Octree & 3D & \starth & \startw & \startw\\
Stixel & 2.5D & \starth & $\star$ & $\star$\\
TSDF & 3D & \startw & \starth & \startw\\
CSG & 3D & \starth & \starth & $\star$\\
\bottomrule
\end{tabular}
\caption{Comparison between data representations. The symbols $\star$, $\star\star$ and $\star\star$$\star$ represent low, moderate and high respectively.  }
\label{tab:data_rep}
\end{table}

\noindent
$\bullet$ \textbf{3D Mesh:}
The mesh representation encodes a 3D object geometry in terms of a combination of edges, vertices, and faces. A mesh that represents the surface of a 3D object using polygon (e.g., triangles or quadrilaterals) shaped faces is termed as the `polygon mesh.' A mesh might contain arbitrary polygons but a `regular mesh' is composed of only a single type of polygons. A commonly used mesh is a triangular mesh that is composed entirely of triangle shaped faces. In contrast to polygonal meshes, `volumetric meshes' represent both the interior volume along with the object surface. 

\noindent
$\bullet$ \textbf{Depth Channel and Encodings:}
A depth channel in a 2.5D representation shows the estimated distance of each pixel from the viewer. This raw data has been used to obtain more meaningful encodings such as HHA \cite{Gupta2014}. Specifically, this geocentric embedding encodes depth image using height above the ground, horizontal disparity and angle with gravity for each pixel. 

\noindent
$\bullet$ \textbf{Octree Representations:}
An octree is a voxelized representation of a 3D shape that provides high compactness. The underlying data structure is a tree where each node has eight children. The idea is to divide 3D occupancy of an object recursively into smaller regions such that empty and similar voxels are represented with bigger voxels. An octree of an object is obtained by a hierarchical process as follows: start by considering 3D object occupancy as a single block, divide it into eight octants. Then, octants that partially contain an object part are further divided. This process continues until a minimum allowed size is reached. The octants can be labeled based on the object occupancy.

\begin{table*}[ht]
\caption{Comparison between various publicly available 2.5/3D indoor datasets. } 
\label{Dataset-table}
\centering
\newlength\q
\setlength\tabcolsep{1.0pt}
\setlength\q{\dimexpr .0865\textwidth }
\scalebox{0.78}{
\begin{tabular}{@{}p{.105\textwidth} >{\centering\arraybackslash}p{\q} >{\centering\arraybackslash}p{\q} >{\centering\arraybackslash}p{\q} >{\centering\arraybackslash}p{\q} >{\centering\arraybackslash}p{\q} >{\centering\arraybackslash}p{\q} >{\centering\arraybackslash}p{\q} >{\centering\arraybackslash}p{\q} >{\centering\arraybackslash}p{\q} >{\centering\arraybackslash}p{\q} >{\centering\arraybackslash}p{\q} >{\centering\arraybackslash}p{\q} >{\centering\arraybackslash}p{\q}@{} }
\toprule
\textbf{Dataset} & \textbf{NYUv2} \cite{NYUDv2} & \textbf{SUN3D} \cite{xiao2013sun3d} & \textbf{SUN RGB-D} \cite{song2015sun} & \textbf{Building Parser} \cite{2017arXiv170201105A} & \textbf{Matterport 3D} \cite{chang2017matterport3d} & \textbf{ScanNet} \cite{dai2017scannet}& \textbf{SUNCG} \cite{song2016ssc} & \textbf{RGBD Object} \cite{lai2011large} & \textbf{SceneNN} \cite{hua2016scenenn} &  \textbf{SceneNet RGB-D} \cite{mccormac2016scenenet} & \textbf{PiGraph} \cite{savva2016pigraphs} & \textbf{TUM} \cite{sturm2012benchmark} & \textbf{Pascal 3D+} \cite{xiang2014beyond}\\ \midrule

Year & 2012 & 2013 & 2015 & 2017 & 2017 & 2017 & 2016 & 2011 & 2016   & 2016 & 2016 & 2012 & 2014 \\ 
Type & Real & Real & Real & Real & Real & Real & Synthetic & Real & Real  & Synthetic & Synthetic & Real & Real\\ 
Total Scans & 464 & 415 & - & 270 & - & 1513 & 45,622 & 900 & 100   & 57 & 63 & 39 & -\\
Labels & 1449 images & 8 scans & 10k images & 70k images & 194k images & 1513 scans & 130k images & 900 scans  & 100 scans  & 5M images & 21 scans$^\ast$ & 39 scans & 24k images\\
Objects/Scenes &  Scene & Scene & Scene & Scene & Scene & Scene & Scene & Object & Scene   & Scene & Scene & Scene & Object\\ 
Scene Classes & 26 & 254 & 47 & 11 & 61 & 707 & 24 & - & -   & 5 & 30 & \xmark & - \\ 
Object Classes & 894 & - & 800 & 13 & 40 & 50 - at least & 84 & 51 & 50 - at least   & 255 & 5 subjects & \xmark & 12\\ 
In/Outdoor & Indoor & Indoor & Indoor & Indoor & Indoor & Indoor & Indoor &Indoor & Indoor   & Indoor & Indoor& Indoor & In+Out\\ 
\midrule
\multicolumn{14}{c}{\textbf{Available Data Types}} \\
\midrule
RGB & \cmark & - & \cmark & \cmark & \cmark & - & - & \cmark & - &   \cmark & \xmark & -& \cmark\\ 
Depth & \cmark & \cmark & \cmark & \cmark & \cmark & - & \cmark & \cmark & \cmark   & \cmark & \xmark & - & \xmark\\ 
Video & \cmark & \cmark & \cmark & \xmark & \xmark & \cmark & \xmark & \cmark & \cmark   & \cmark & \cmark & \cmark & \xmark \\ 
Point cloud & \xmark & \xmark & \cmark & \cmark & \cmark & \xmark & \xmark & \xmark & \xmark & \xmark & - & \xmark & \xmark \\ 
Mesh/CAD & \xmark & \xmark & \xmark & \cmark & \cmark & \cmark & \cmark & \xmark & \cmark   & \xmark & - & \xmark & \cmark\\
\midrule
\multicolumn{14}{c}{\textbf{Available Annotation Types}} \\
\midrule
Scene Classes  & \cmark & \xmark & \cmark & \cmark& \cmark & \xmark& \xmark &\xmark & \cmark & \cmark& \xmark& \xmark& \xmark\\
Semantic Label & \cmark & \cmark & \cmark & \cmark& \cmark& \cmark& \cmark & \xmark& \xmark& \cmark& \xmark& \xmark& \xmark\\
Object BB & \cmark & \cmark& \cmark & \cmark & \xmark& \cmark& \cmark& \cmark& \cmark& \cmark& \xmark& \xmark& \cmark\\
Camera Poses & \cmark & \cmark& \xmark& \cmark& \cmark& \cmark& \cmark& \cmark& \cmark& \cmark& \xmark& \cmark& \cmark\\
Object Poses & \cmark & \xmark& \cmark& \xmark& \xmark& \xmark& \xmark& \cmark& \cmark& \xmark& \xmark& \xmark& \cmark\\
Trajectory & \xmark& \xmark& \xmark& \xmark& \xmark& \xmark& \xmark& \xmark& \xmark& \cmark& \xmark& \cmark& \xmark\\
Action& \xmark& \xmark& \xmark& \xmark& \xmark& \xmark& \xmark& \xmark&\xmark& \xmark& \cmark& \xmark& \xmark \\
\bottomrule
\end{tabular}}\vspace{0.2em}
-: means information not available, $\ast$: Average reported; 4.9 actions annotated per scan and there are 298 actions with 8.4s length available.
\end{table*}

\noindent
$\bullet$ \textbf{Stixels:}
The idea of stixels is to reduce the gap between pixel and object level information, thus reducing  the number of pixels in a scene to few hundreds \cite{badino2009stixel}. In stixel representation, a 3D scene is represented by vertically oriented rectangles with a certain height. Such a representation is specifically useful for traffic scenes, but limited in its capability to encode generic 3D scenes.

\noindent
$\bullet$ \textbf{Truncated Signed Distance Function:}
Truncated signed distance function (TSDF) is another volumetric representation of a 3D scene. Instead of mapping a voxel to 0 or 1, each voxel in the 3D grid is mapped to the signed distance to the nearest surface. The signed distance is negative if the voxel lies with in the shape and positive otherwise. RGB-D camera (e.g., Kinect) representations are based on TSDF further fuse them to obtain a complete 3D model.

\noindent
$\bullet$ \textbf{Constructive Solid Geometry:}
Constructive solid geometry (CSG) is a building block technique in which simple objects such as cubes, spheres, cones, and cylinders are combined with a set of operations such as union, intersection, addition, and subtraction to model complex objects. CSG is represented as a binary tree with primitive shapes and the combination operations as its nodes. This representation is often used for CAD models in computer vision and graphics. 

\section{Datasets}
High quality datasets play important role in development of machine vision algorithms. Here, we review important datasets, Table \ref{Dataset-table},  for scene understanding available to researchers. 

\noindent
$\bullet$ \textbf{NYU-Depth:}
Silberman et al. introduced NYU Depth v1 \cite{NYUDv1} and v2 \cite{NYUDv2} in 2011 and 2012, respectively. NYU Depth v1 \cite{NYUDv1} consists of 64 different indoor scenes with 7 scene types. There are 2347 RGBD images available. The dataset is roughly divided into 60\%/40\% for train/test respectively. NYU Depth v2 \cite{NYUDv2} consists of 1449 RGBD images representing 464 different indoor scenes with 26 scene types. Pixel level labeling is provided for each image. There are 795 images in train set and 654 images in the test set. Both versions were collected using Microsoft Kinect. 

\noindent
$\bullet$ \textbf{Sun3D:}
Sun3D \cite{xiao2013sun3d} dataset provides videos of indoor scenes that are registered into point clouds. The semantic class and instance labels are automatically propagated through the video from the seed frames. Dataset provides 8 annotated sequences, and there are in total of 415 sequences available for 254 different spaces in 41 different buildings.

\noindent
$\bullet$ \textbf{SUN RGB-D:}
Sun RGB-D \cite{song2015sun} contains 10335 indoor images with dense annotations in 2D and 3D for both objects and indoor scenes. It includes 146617 2D polygons and 64595 3D bounding boxes for object orientation as well as scene category and 3D room layout for each image. There are 47 scene categories, 800 object categories, and each image contains on average 14.2 objects. This data set is captured by four different kinds of RGB-D sensors and designed to evaluate scene classification, semantic segmentation, 3D object detection, object orientation, room layout estimation and total scene understanding. The data is divided into training and test sets such that each sensor data has half allocation for training and the other half for testing.

\noindent
$\bullet$ \textbf{Building Parser:}
Armeni et al.\cite{2017arXiv170201105A} provide a dataset with instance level semantic and geometric annotations. The dataset was collected from 6 different areas, and it contains 70496 RGB and 1412 equirectangular RGB with their corresponding depths, semantic annotations, surface normal, global XYZ openEXR format and camera metadata. These 6 different areas are divided into training and test splits with a 3-fold cross-validation scheme, i.e., training with 5 areas, training with 4 areas and training with 3 areas while testing with the rest of scans in each case.

\noindent
$\bullet$ \textbf{ScanNet:}
ScanNet \cite{dai2017scannet} is the 3D reconstructed dataset with 2.5 million data frames obtained from 1513 RGB scans. These 1513 annotated scans represent 707 different spaces including small ones like closets, bathrooms, and utility rooms, and large spaces like classrooms, apartments, and libraries. These scans are annotated with instance level semantic category labels. There are 1205 scans in the training set and another 312 scans in the test set.

\noindent
$\bullet$ \textbf{PiGraph:}
Savva et al. \cite{savva2016pigraphs} proposed the PiGraph representation to link human poses with object arrangements in indoor environment. Their dataset contains 30 scenes and 63 video recordings for five human subjects obtained by Kinect v2. There are 298 actions available in approximately 2-hour of recordings. Each recording is about 2 minute long with on average 4.9 action annotations. They link 13 common human actions like sitting, reading to 19 object categories such as couch and computer monitor.

\noindent
$\bullet$ \textbf{SUNCG:}
SUNCG \cite{song2016ssc} is a densely annotated, large scale dataset of 3D scenes. It contains 45622 different scenes that are semantically annotated at object level. These scenes are manually created using the Planner5D platform \cite{Planner5D}. Planner5D is an interior design tool that can be used to generate novel scene layouts. This dataset contains around 49K floor maps, 404K rooms and 5697K object instances covering 84 object categories. All the objects are manually assigned to a category label. 

\noindent
$\bullet$ \textbf{PASCAL3D+:}
Xiang et al. \cite{xiang2014beyond} introduced Pascal3D+ for 3D object detection and pose estimation tasks. They picked 12 categories of objects including airplane, bicycle, boat, bottle, bus, car, chair, motorbike, dining table, sofa, tv monitor, and train (from Pascal VOC dataset \cite{everingham2010pascal}) and performed 3D labeling. Further, they included additional images for each category from ImageNet dataset \cite{imagenet2009}. The resulting dataset has around 3000 object instances per category.

\noindent
$\bullet$ \textbf{RGBD Object:}
This dataset \cite{lai2011large} provides video recordings of 300 household objects assigned to 51 different categories. The objects are categorized using WordNet hypernym-hyponym relationships. There are 3 video sequences for each object category recorded by mounting Kinect camera at different heights. The videos are recorded at a frame rate of 30Hz with 640$\times$480 resolution for RGB and depth images. This dataset also contains 8 annotated video sequences of indoor scene environments. 

\noindent
$\bullet$ \textbf{TUM:}
\cite{sturm2012benchmark} provides a large-scale dataset for tasks like visual odometry and SLAM (simultaneous localization and mapping). The dataset contains RGB and depth images obtained using Kinect sensor along with the groundtruth sensor trajectory (poses and positions). The dataset is recorded at a frame rate of 30Hz with 640$\times$480 resolution for RGB and depth images. The groundtruth trajectory was obtained using high speed cameras working at 100Hz. There are in total 39 sequences of indoor environments.

\noindent
$\bullet$ \textbf{SceneNN:}
SceneNN \cite{hua2016scenenn} is a fine-grain annotated RGBD dataset of indoor environments. It consists of 100 scenes where each scene is represented as a triangular mesh having per vertex and per pixel annotations. The dataset is further enriched by providing information such as oriented bounding boxes, axis-aligned bounding boxes and object poses.


\noindent
$\bullet$ \textbf{Matterport3D:}
Matterport3D \cite{chang2017matterport3d} provides a diverse and large-scale RGBD dataset for indoor environments. This dataset provides 10800 panoramic images covering $360^{\circ}$ views captured by Matterport camera. Matterport camera comes with three color and three depth cameras. To get a panoramic view, they rotated the Matterport camera by $360^{\circ}$, stopping at six locations and capturing three RGB images at each location. The depth cameras continuously acquired depth information during rotation, which was then aligned with each color image. Each panoramic image contains 18 RGB images. In total, there are 194400 color and depth images representing indoor scenes of 90 buildings. This dataset is annotated for 2D and 3D semantic segmentations, camera poses and surface reconstructions.  

\noindent
$\bullet$ \textbf{SceneNet RGB-D:}
SceneNet is a synthetic video dataset which provides pixel level annotations for nearly 5M frames. The dataset is divided into training set with 5M images while validation and test set contains 300K images. This dataset can be used for multiple scene understanding tasks including semantic segmentation, instance segmentation and object detection.



\section{Core Techniques}\label{sec:core}

\begin{figure}[!htp]
\centering
\includegraphics[width=\columnwidth, height=5cm]{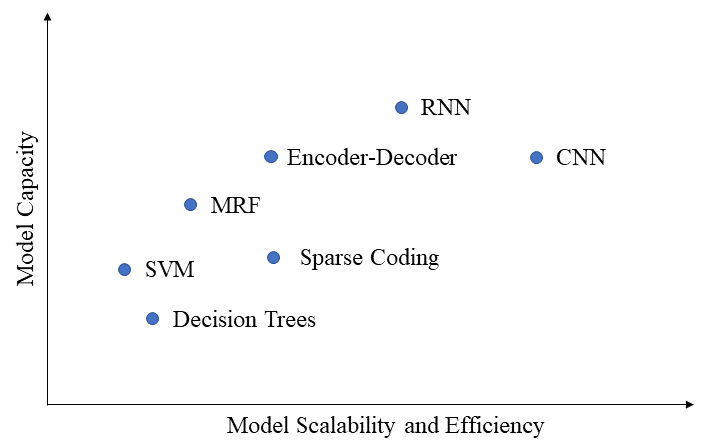}
\caption{Core-Techniques Comparison.}
\label{fig:methods}
\end{figure}

We begin with an overview of the core techniques employed in the literature for various scene understanding problems. For each respective technique, we discuss its pros and cons in comparison to other competing methods (Fig.~\ref{fig:methods}). Later in this survey, we provide a detailed description of recent methods that built on the strengths of these core techniques or attempt to resolve some of their weaknesses. In this regard, several hybrid approaches have also been proposed in the literature e.g., \cite{badrinarayanan2015segnet, socher2012convolutional, zheng2015conditional}, which combine strengths of different core techniques to achieve better performances.   

\subsection{Convolutional Neural Networks}
\begin{figure}[h]
\centering
\includegraphics[width=\columnwidth]{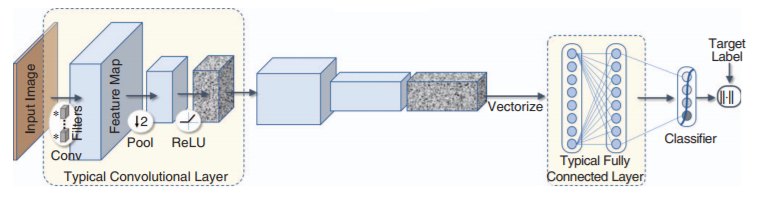}
\caption{A basic CNN architecture with a convolution, pooling, activation along with a fully connected layer.}
\label{fig:CNN}
\end{figure}

An artificial neural network (ANN) consists of a number of computational units, which are arranged in multiple interconnected layers. Convolutional Neural Network (CNN) is a special type of ANN whose main building blocks consist of filters that are spatially convolved with the inputs to generate output feature maps. A building block is called as the `convolutional layer,' which usually repeats several times in a CNN architecture. A convolution layer drastically reduces the network parameters through weight-sharing. Further, it makes the network invariant to translations in the input domain. The convolution layers are interleaved with other layers such as pooling (to subsampling inputs), normalization (to rescale activations) and fully connected layers (to reduce the feature dimensions or to densely connect input and output units). A simple CNN architecture is illustrated in Fig.~\ref{fig:CNN}, which shows the above-mentioned layers. 

CNNs have shown excellent performance on many scene understanding tasks (e.g., \cite{girshick2015fast, simonyan2014very, qi2016pointnet, zhang2014panocontext, zhang2016deepcontext, li2016fpnn, su2015multi}). Some distinguishing features that permit CNNs to achieve superior results can be considered as end-to-end learning of network weights, scalability to large problem sets and computationally efficiency in their derivation of large-scale models. However, it is nontrivial to incorporate prior knowledge and rich relationships between variables in a traditional CNN. Besides, conventional CNNs do not operate on arbitrarily shaped inputs such as point clouds, meshes, and variable length sequences.

\subsection{Recurrent Neural Networks}
\begin{figure}[h]
\centering
\includegraphics[width=\columnwidth]{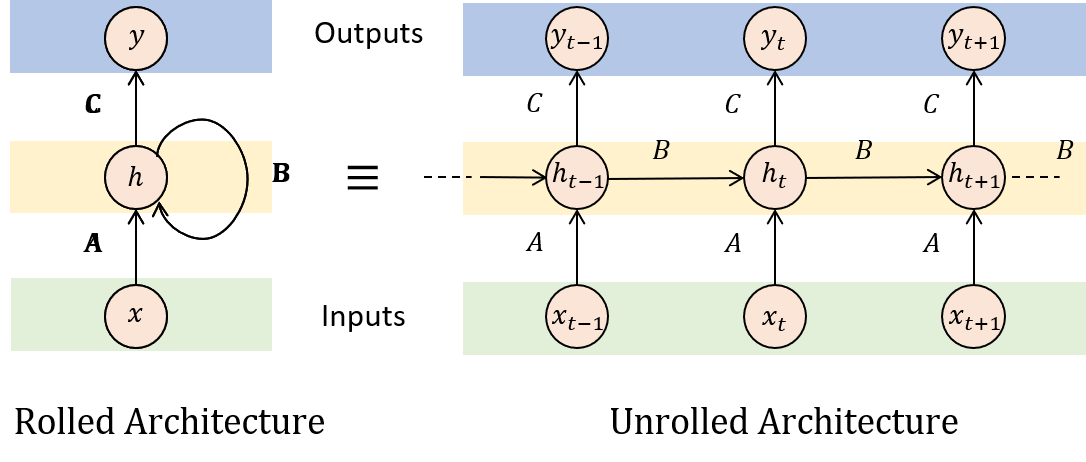}
\caption{A basic RNN architecture in the rolled and unrolled form. }
\label{fig:RNN}
\end{figure}

While CNNs are feedforward networks (i.e., they do not have cycles or loops), Recurrent Neural Network (RNN) has a feedback architecture where information flow happens along directed cycles. This capability allows them to work with arbitrary sized inputs and outputs. RNNs exhibit memorization ability and can store information and sequence relationships in their internal memory states. A prediction at a specific time instance `$t$' can then be made while considering the current input as well as the previous hidden states (Fig.~\ref{fig:RNN}). Similar to the case of convolution layer in CNNs where weights are shared along the spatial dimensions of the inputs, the RNN weights are shared along the temporal domain, i.e., same weights are applied to inputs at each time instance. Compared to CNNs, RNNs have considerably less number of parameters due to such weight sharing mechanism.

As discussed above, the hidden state of the RNN provides a memory mechanism, but it is not effective when the goal is to remember long-term relationships in sequential data. Therefore, RNN only accommodates short-term memory and faces with difficulties in `remembering' (a few time-steps away) old information processed through it. To overcome this limitation, improved versions of recurrent networks have been introduced in the literature which include the Long Short-Term Memory (LSTM) \cite{hochreiter1997long}, Gated Recurrent Unit (GRU) \cite{cho2014properties}, Bidirectional RNN (B-RNN) \cite{graves2005framewise} and Neural Turing Machines (NTM) \cite{graves2014neural}. These architectures introduce additional gates and recurrent connections to improve the storage ability. Some representative works in 3D scene understanding that leverage the strengths of RNNs include \cite{socher2012convolutional,choy20163d}. 

\subsection{Encoder-Decoder Architectures}
\begin{figure}[!h]
\centering
\includegraphics[width=\columnwidth]{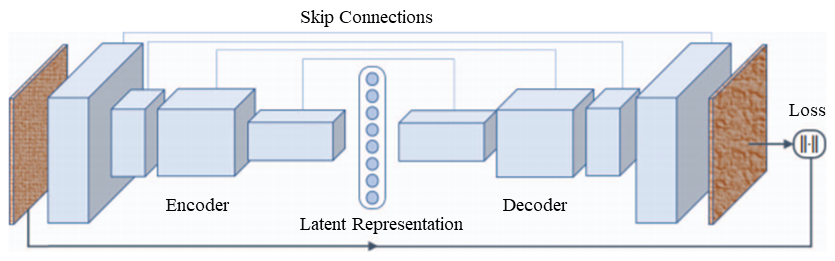}
\caption{A basic autoencoder architecture.}
\label{fig:Encoder-Decoder}
\end{figure}

The encoder-decoder networks are a type of ANNs, which can be used for both supervised and unsupervised learning tasks. Given an input, an `encoder' module learns a compact representation of the data which is then used to reconstruct either the original input or an output of another form (e.g., pixel labels for an image) using a `decoder' (Fig.~\ref{fig:Encoder-Decoder}). This type of network is called an autoencoder when  the input to the encoder is reconstructed back using the decoder. Autoencoders are typically used for unsupervised learning tasks. A closely related variant of an autoencoder is a variational autoencoder (VAE) that introduces constraints on the latent representation learned by the encoder. 

Encoder-decoder style networks have been used in combination with both convolutional \cite{badrinarayanan2015segnet} and recurrent designs \cite{xu2015show}. The applications of such designs for scene understanding tasks include \cite{badrinarayanan2015segnet, kehl2016deep, kendall2015bayesian, Riegler2017OctNet, nguyen2016detecting, dai2016shape}. The strength of these approaches is to learn a highly compact latent representation from the data, which is useful for dimensionality reduction and can be directly employed as discriminative features or transformed using a decoder to generate desired outputs. In some cases, the encoding step leads to irreversible loss of information which makes it challenging to reach the desired output. 

\subsection{Markov Random Field} 

Markov Random Field (MRF) is a class of undirected probabilistic models that are defined over arbitrary graphs. The graph structure is composed of nodes (e.g., individual pixels or super-pixels in an image) interconnected by a set of edges (connections between pixels in an image). Each node represents a random variable which satisfies Markovian property, i.e., conditional independence from all variables if the neighboring variables are known. The learning process for  a MRF involves estimating a generative model i.e., the joint probability distribution over input (data; $\mathbf{X}$) and output (prediction; $\mathbf{Y}$) variables i.e., $P(\mathbf{X}, \mathbf{Y})$. For several problems, such as classification and regression, it is more convenient to directly model the conditional distribution $P(\mathbf{Y}|\mathbf{X})$ using the training data. The resulting discriminative Conditional Random Field (CRF) models often provide more accurate predictions. 

Both MRF and CRF models are ideally suited for structured prediction tasks where the predicted outputs have inter-dependent patterns instead of, e.g., a single category label in the case of classification. Scene understanding tasks  often involve structured prediction e.g., \cite{lin2013holistic,wang2015holistic,khan2015separating, kim20133d, jiang2016modeling, koppula2016anticipating}. These models allow the incorporation of context while making local predictions. The context can be encoded in the model by pair-wise potentials and clique potentials (defined over groups of random variables). This results in more informed and coherent predictions which respect the mutual relationships between labels in the output prediction space. However, training and inference in several of such model instantiations are not tractable, which makes their application challenging.

\subsection{Sparse Coding}

Sparse coding is an unsupervised method used to find a set of basis vectors such that an input vector `$\bf x$' can be represented by their linear sparse combination \cite{lee2007efficient}. The set of basis vectors is called as a `dictionary' ($\bf D$), which is typically learned over the training data. Given the dictionary, a sparse vector $\bf \alpha$ is calculated such that the input $\bf x$ can be accurately reconstructed back using $\bf D$ and $\bf \alpha$. Sparse coding can be seen as decomposing a non-linear input into sparse combination of linear vectors. If the basis vectors are large in number or when the dimension of feature vectors is high, optimization process required to calculate $\bf D$ and $\bf \alpha$ can be computationally expensive. Examples of sparse coding based approaches in scene understanding literature include \cite{myers2015affordance,bo2014learning,wang2013learning}.  

\begin{figure}[t]
\centering
\includegraphics[width=\columnwidth, height=4cm]{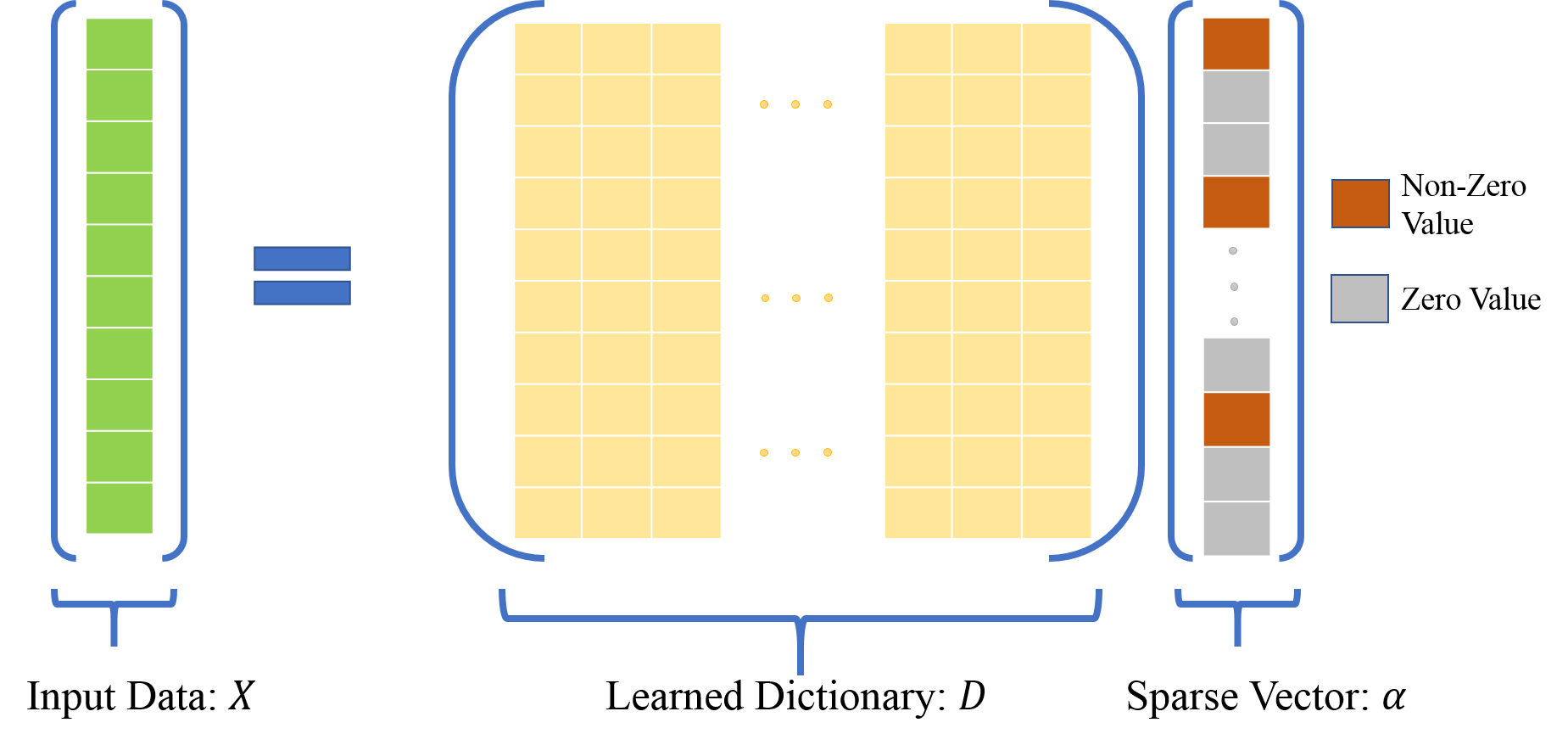}
\caption{Dictionary learning for sparse coding.}
\label{fig:sparse-coding}
\end{figure}

\subsection{Decision Forests}
A decision tree is a supervised algorithm that classifies data based on a graph based hierarchy of rules learned over the training set. Each internal node in a decision tree represents a test or attribute (true or false question) while each leaf node represents the decision on a class label. To build a decision tree, we start with a root node that receives all the training data and based on the test question we split the data into subsets. These subsets then become the inputs for the next two child nodes. This process continues until we produce the best possible distribution of the labels at each node, i.e., a total unmixing of data is achieved. One can quantify the mixing or uncertainty at a single node by a metric called `Gini impurity' which can be minimized by devising rules based on information gain. We can use these measures to ask the best question at each node and continue to build the decision tree recursively until there are no more questions to ask. Decision trees can quickly overfit the training data which can be rectified by using random forests \cite{breiman2001random}. Random forest builds an ensemble of decision trees using a random selection of data and produces class labels based on many decisions trees. Representative works using random forests include \cite{tejani2014latent, mottaghi2015coarse,firman2016structured, myers2015affordance, bonde2014robust}.
\subsection{Support Vector Machines}
When it comes to classifying the n-dimensional data points, the goal is not only to separate the data into a certain number of categories but to find such a dividing boundary that offers maximum possible separation between classes. Support vector machine (SVM) \cite{hearst1998support,gunn1998support} offers such a solution. SVM is a supervised method that separates the data with a linear hyperplane, also called maximum-margin hyperplane, that offers maximum separation between any combination of two classes. SVM can also be used to learn nonlinear classification boundaries with the kernel trick. The idea of kernel trick is to project the nonlinearly separable low dimensional data into a high dimensional space where the data is linearly separable. After applying SVM into high dimensional linearly separable space, project the solution back to low-dimensional, nonlinearly separable space to get a nonlinear hyperplane boundary.
\vspace{1cm}
\section{A Taxonomy of Problems}\label{sec:sub-problems}
\subsection{Image Classification }
\subsubsection{Prologue and Significance}
Image recognition is a basic, yet one of the fundamental tasks for visual scene understanding. Information about the scene or object category can help in more sophisticated tasks such as scene segmentation and object detection. Classification algorithms are being used in diverse areas such as medical imaging, self-driving cars and context-aware devices. In this section, we will provide an overview of some of the most important methods for 2.5D/3D scene classification. These approaches employ a diverse set strategies including handcrafted features \cite{gupta2013perceptual}, automatic feature learning \cite{socher2012convolutional,su2015multi}, unsupervised learning \cite{wu2016learning} and work on different 3D representations such as voxels \cite{qi2016volumetric} and point clouds \cite{qi2016pointnet}.

\subsubsection{Challenges}
Important challenges for image classification include:
\begin{itemize}
\item 2.5/3D data can be represented in multiple ways as discussed above. Challenge then is to choose the data representation that provides maximum information with minimum computational complexity. 
\item A key challenge is to distinguish between fine-grained categories and appropriately model intra-class variations.
\item Designing algorithms that can handle illuminations, background clutter and 3D deformations.
\item Designing algorithm that can learn from limited data.
\end{itemize}
\subsubsection{Methods overview}

A bottom-up approach to scene recognition was introduced in \cite{gupta2013perceptual}, where the constituent objects were first identified to improve the scene classification accuracy. In this regard, they first extended a contour detection method (gPb-ucm \cite{arbelaez2011contour}) to RGB-D images by effectively incorporating the depth information. Note that the gPb-ucm approach produces hierarchical image segmentation by using contour information \cite{arbelaez2011contour}.  The predicted semantic segmentation maps were used as features for scene classification. They used a special pyramid formulation, similar to spatial pyramid matching approach \cite{lazebnik2006beyond}, along with SVM as a classifier.

Sochar et al. \cite{socher2012convolutional} introduced a method to learn features from RGB-D images using RNN. They used a convolutional layer to learn low-level features which were then passed through multiple RNNs to learn high-level feature representations before feeding to a classifier. At the CNN stage, RGB and depth patches were clustered using k-means to obtain the convolutional filters in an unsupervised manner \cite{coates2011analysis}. These filters were then convolved with images to get low-level features. After performing dimensionality reduction via pooling process, these features were then fed to multiple RNNs which recursively operate in a tree-like structure to learn high-level feature representations. The outputs of these multiple RNNs were concatenated to form a final vector which is forwarded to a SoftMax classifier for the final decision. An interesting insight of their work is that weights of RNNs were not learned through back propagation rather set to random values. Increasing the number of RNNs resulted in an improved model classification accuracy. Another important insight of their work is that RGB and depth images produce independent complimentary features and their combination improves the model accuracy. Similar to this work, \cite{eitel2015multimodal} extracted features from RGB and depth modalities via two stream networks, which were then fused together. \cite{wang2016modality} extended the same pattern by learning features from modalities like RGB, depth and surface normals. They also proposed to encode local CNN features with fisher vector embedding and then combine them with global CNN features to obtain better representations.

In an effort to build a 3D shape classifier, Wu et al. \cite{wu20153d} introduced a convolutional deep belief network (DBN) trained on 3D voxelized representations. Note that different from restricted Boltzmann machines (RBM), a DBN is a directed model that can detect patterns from unlabeled data. An RBM is a two-way translator that takes input in a forward pass and translate it to latent representation that encodes the input, while in the backward pass it takes the latent representation and translates it back to reconstruct the input. \cite{hinton2006fast} showed that DBN could learn the joint distributions of 2D image pixels and labels. \cite{wu20153d} extended this idea to learn joint probabilistic distributions of 3D voxels and object categories. The novelty in their architecture is to introduce convolutional layers which, in contrast to fully connected layers, allow weight sharing and significantly reduce the number of parameters in the DBN. On similar lines, \cite{maturana2015voxnet} advocates using 3D CNN on a voxel grid to extract meaningful representations while \cite{li2016fpnn} proposes to approximate 3D spaces as volumetric fields to deal with the computational cost of directly applying 3D CNN to voxels.

\begin{figure}[!htp]
 \centering
\includegraphics[width=\linewidth]{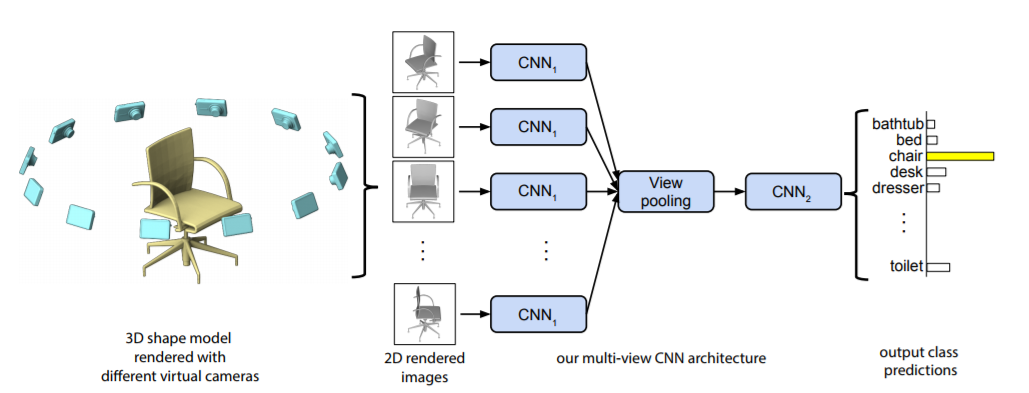}
\caption{Multi-view CNN for 3D shape recognition \cite{su2015multi}. Extracted features of different views from CNN$_1$ are pooled together before passing through CNN$_2$ for final score prediction. (Courtesy of \cite{su2015multi})}
 \label{mvcnn}
\end{figure}

Though it seems logical to build a model that can directly consume 3D shapes to recognize them (e.g., \cite{wu20153d}), however the 3D resolution of a shape must be significantly reduced to allow feasible training of a deep neural network. As an example, 3D ShapeNets \cite{wu20153d} used a $30\times 30\times 30$ binary voxel grid to represent 3D shapes. Su et al. \cite{su2015multi} provided evidence that 3D shapes can be recognized by their 2D views and presented a multi-view CNN (MVCNN) architecture to recognize 3D shapes that can be trained on 2D rendered views. They used the Phong reflection method \cite{phong1975illumination} to render 2D views of 3D shapes. Afterwards,  a pre-trained VGG-M network \cite{chatfield2014return} was fine-tuned on these rendered views. To aggregate the complementary information across different views, each rendered view was passed through the first part of the network (CNN$_1$) separately, and the results across views were combined using element-wise maximum operation at the pooling layer before passing them through the rest of network (CNN$_2$, see Figure \ref{mvcnn}). MVCNN thus combines the multiple view information to better recognize 3D shapes. While MVCNN represented 3D shapes with multiple 2D images, \cite{shi2015deeppano} proposes to convert 3D shapes into a panoramic view. 

We have observed so far that the existing models utilize different 3D shape representations (i.e., volumetric, multi-view and panoramic) to extract useful features. Intuitively volumetric representations should contain more information about the 3D shape, but multi-view CNNs \cite{su2015multi} perform better than volumetric CNNs \cite{wu20153d}. Qi et al. \cite{qi2016volumetric} argued that network architecture differences and input resolutions are the reasons for the gap in performance. Inspired from multi-view CNNs, \cite{qi2016volumetric} introduced a multi-orientation network architecture that takes various orientations of input voxel grid, extract features for each orientation using a shared network CNN$_1$, pooled the features before passing through CNN$_2$. To take benefit of well trained 2D CNN, they introduced 3D-to-2D projection using anisotropic probing kernels to classify the 2D projection of the 3D shape. They also improved multi-view CNNs \cite{su2015multi} performance by introducing a multi-resolution scheme. Inspired by the performance efficiency of MVCNN and volumetric CNN, \cite{hegde2016fusionnet} fused both modalities to learn better features for classification.

\begin{figure}[!htp]
 \centering
\includegraphics[width=\linewidth]{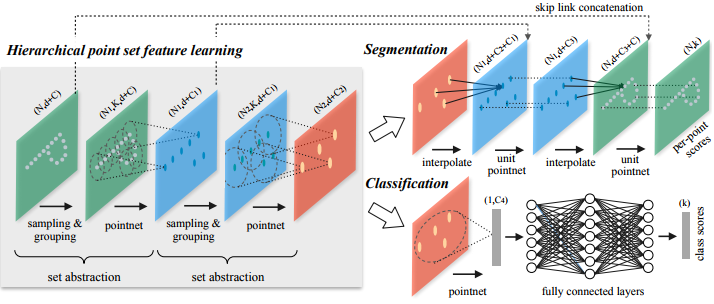}
\caption{PointNet++ \cite{qi2017pointnet++} architecture for point cloud classification and segmentation. PointNet architecture \cite{qi2016pointnet} is being used in a hierarchical fashion to extract local geometric features (Courtesy of \cite{qi2017pointnet++}).}
 \label{pointnet++}
\end{figure}

A point cloud is a primary geometric representation captured by 3D scanners. However, due to its variable number of points from one shape to another, it needs to be transformed to a regular input data format, e.g., voxel grid or multi-view images. This transformation, however, can increase the data size and result in undesired artifacts. PointNet \cite{qi2016pointnet} is a deep net architecture that can consume point clouds directly and output the class label. PointNet takes a set of points as input, performs feature transformations for each point, assemble feature across points via max-pooling and output the classification score. Even though PointNet process unordered point clouds but by design, it lacks the ability to capture local contextual features due to metric space of the points. Just like a CNN architecture learns hierarchical features mapped from local patterns to more abstract motifs, Qi et al. \cite{qi2017pointnet++} applied PointNet on point sets recursively to learn local geometric features and then grouped these features to produce high-level features for the whole point set (see Figure \ref{pointnet++}). A similar idea was adopted in \cite{zeng20173dcontextnet}, which performed a hierarchical feature learning over a k-d tree structured partitioning of 3D point clouds. 

Finally, we would like to describe an unsupervised method for 3D object recognition. By combining the power of volumetric CNN \cite{wu20153d} and generative adversarial networks (GAN) \cite{goodfellow2014generative}, Wu et al. \cite{wu2016learning} presented a novel framework called 3D-GAN for 3D object generation and recognition. An adversarial discriminator in GAN learns to classify whether an object is real or synthesized. \cite{wu2016learning} showed that representations learned by an adversarial discriminator without supervision could be used as features for linear SVM to obtain classification scores of 3D objects.

\subsection{Object Detection}
\subsubsection{Prologue and Significance}
Object detection deals with recognizing object instances and categories. Usually, an object detection algorithm outputs both the location (defined by a 2/3D bounding box around the visible parts of an object instance) and the class of an object, e.g., sofa, chair. This task has high significance for applications such as self-driving cars, augmented and virtual reality. However, in applications such as robot navigation, we need so-called `amodal object detection' that tries to find an object's location as well as its complete shape and orientation in 3D space when only a part of it is visible. In this section, we review 2.5/3D object detection methods mainly focused on indoor scenes. We observe the role of handcrafted features in object recognition \cite{khan2015separating, jiang2014finding, song2014sliding} and the recent transition to deep neural networks based region proposal (object candidate) generation and object detection pipelines \cite{chen20153d,song2016deep, deng2017amodal}. Apart from the supervised models, we also review unsupervised 3D object detection techniques \cite{novotny2017learning}. 

\subsubsection{Challenges}
Key challenges for object detection are as follows:
\begin{itemize}
\item Real world environments can be highly cluttered and object identification in such environments is very challenging.
\item Detection algorithm should also be able to handle view-point and illuminations variations and deformations.
\item In many scenarios, it is necessary to understand the scene context to successfully detect objects. 
\item Objects categories have a long-tail (imbalanced) distribution, which makes it challenging to model the infrequent classes.  
\end{itemize}
 
\subsubsection{Methods Overview}
Jiang et al. \cite{jiang2013linear} proposed a bottom-up approach to detect 3D object bounding boxes using RGB-D images. Starting from a large number of object proposals, physically plausible boxes were identified by using volumetric properties such as solidness, 3D overlap, and occlusion relationships. Later, \cite{jiang2014finding} argued that convex shapes are more descriptive than cuboids and can be used to represent generic objects. A limitation of these techniques is that they ignore semantics in a scene and are limited to finding object shape. In a real-world scenario, a scene can contain regular objects (e.g., furniture) as well as cluttered regions (e.g., clothes pile on a bed). Khan et al. \cite{khan2015separating} extended the technique presented in \cite{jiang2013linear} to jointly detect 3D object cuboids and indoor structures (e.g., floor, walls) along with pixel level labeling of cluttered regions in RGB-D images. A CRF model was used to model the relationships between objects and cluttered regions in indoor scenes. However, these approaches do not provide object-level semantic information apart from a broad categorization into regular objects, clutter, and background.

Several object detection approaches \cite{song2014sliding,zia2015towards,song2016deep,  bo2014learning} have been proposed to provide category information and location of each detected object instance. \cite{bo2014learning} proposed a sparse coding network to learn hierarchical features for object recognition from RGB-D images. Sparse coding models data as a linear combination of atoms belonging to a codebook subject to sparsity constraints. The multi-layer network \cite{bo2014learning} learns codebooks for RGB-D images via K-SVD algorithm \cite{aharon2006rm}, using the grayscale, color, depth, and surface normal information. The feature hierarchy is built as the receptive field size increases along the network depth which helps to learn more abstract representations of RGB-D images. They used orthogonal matching pursuit \cite{pati1993orthogonal}  algorithm for sparse coding, feature pooling to reduce dimensionality and contrast normalization at each layer of the network.

The performance of an object detection algorithm can suffer due to variations in object shapes, viewpoints, illumination, texture, and occlusion. Song et al. \cite{song2014sliding} proposed a method to deal with these variations by exploiting synthetic depth data. They take a collection of 3D CAD models of an object category and render it from different viewpoints to obtain depth maps.  The feature vectors corresponding to depth maps of an object category are then used as positives to train exemplar SVM \cite{malisiewicz2011ensemble} against negatives obtain from RGB-D datasets \cite{NYUDv2}.  At test time, a 3D window is slid on the scene to be classified by the learned SVMs. While \cite{song2014sliding} represented objects with CAD models, other representations are also explored in literature such as \cite{zia2015towards,ren2016three} proposed 3D deformable wire-frame modeling and cloud of oriented gradients representation, respectively, and \cite{karpathy2013object} build object detector based on 3D mesh representation of indoor scenes. 

\begin{figure}[!t]
\centering
 \subfloat[3D Region Proposals Network.]{\label{figur:51}\includegraphics[width=\linewidth, height=4cm]{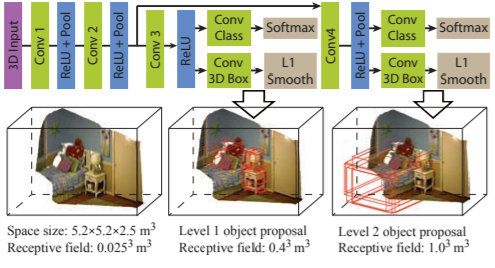}}\\
 \subfloat[Object Detection and 3D box regression Network.]
 {\label{figur:52}\includegraphics[width=\linewidth, height=3cm]{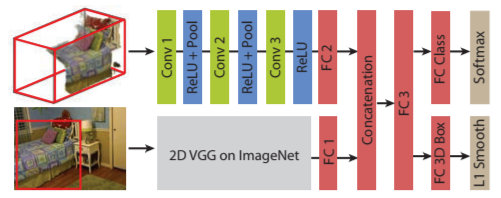}}
\caption{(a) 3D region proposals extraction using using CNNs operating on 3D volumes. (b) A combination of 2D and 3D CNN jointly used to predict object category and location through regression. (Courtesy of \cite{song2016deep})}
\label{amode_ob}
\end{figure}

Typically, an object detection algorithm produces a bounding box on visible parts of the object on an image plane, but for practical reasons, it is desirable to capture the full extent of the object regardless of occlusion or truncation. Song et al. \cite{song2016deep} introduced a deep learning framework for amodal object detection. They used three deep network architectures to produce object category labels along with 3D bounding boxes. First, a 3D network called Region Proposal Network (RPN) takes a 3D volume generated from depth map and produces 3D regional proposals for the whole object.  Each region proposal is feed into another 3D convolutional net, and its 2D projection is fed to a 2D convolutional network to jointly learn color and depth features. The final output is the object category along with the 3D bounding box (see Figure \ref{amode_ob}). A limitation of this work is that the object orientation is not explicitly considered. As \cite{sedaghat2016orientation} demonstrated with their oriented-boosted 3D CNN (Vox-net), this can adversely affect the detection performance and joint reasoning about the object category, location, and 3D pose leads to a better performance.

\begin{figure*}[!t]
\centering
\includegraphics[width=\linewidth]{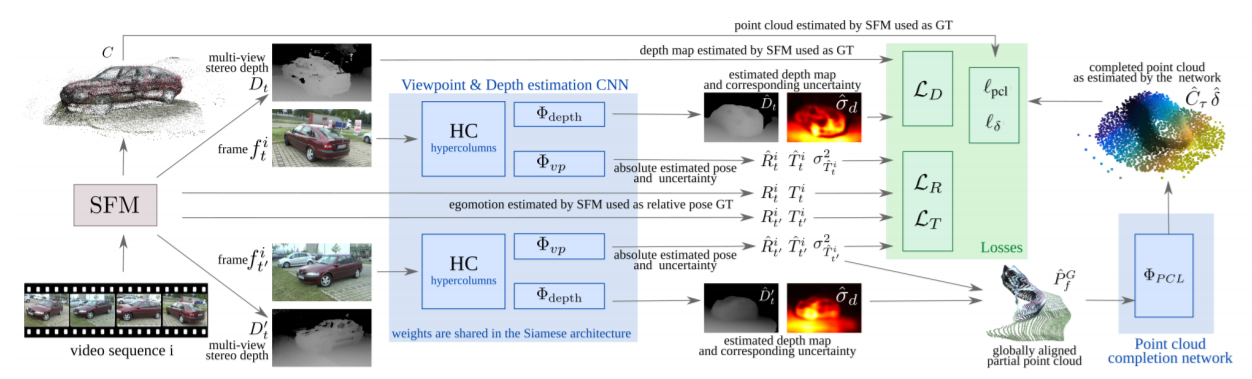}
\caption{Siamese factorization network \cite{novotny2017learning} that takes pair of frames and estimate view point, depth and finally produce point cloud of estimated 3D geometry. Once the network is trained, it can produce viewpoint, depth and 3D geometry from single image at test time. (Courtesy of \cite{novotny2017learning}) }
\label{siamese}
\end{figure*}

Deng et al. \cite{deng2017amodal} introduced a novel neural network architecture based on Fast-RCNN \cite{girshick2015fast} for 3D amodal object detection. Given RGB-D images, they first computed the 2D bounding boxes using multiscale combinatorial grouping (MCG) \cite{Arbelaez2014} over superpixel segmentations. For each 2D bounding box, they initialized the location of the 3D box. The goal is then to predict the class label and adjust the location, orientation, and dimension of the initialized 3D box. In doing so, they successfully showed the correlation between 2.5D features and 3D object detections. Novotny et al. \cite{novotny2017learning} proposed to learn 3D object categories from videos in an unsupervised manner. They used Siamese factorization network architecture to align videos of 3D objects to estimate viewpoint, then produce depth maps using the estimated viewpoints, and finally, the 3D object model is constructed using the estimated depth map (see Figure \ref{siamese}).

Finally, we would like to mention 3D object detection with attention mechanism. To understand a specific aspect of an image, humans can selectively focus their attention on a specific part of the image to gain information. Inspired by this, \cite{xu20163d} proposed 3D attention model that scan a scene to select best views and focus on most informative regions for object recognition task. It further combines the 3D CAD models to replace the actual objects, such that a full 3D scene can be reconstructed. This demonstrates how object detection can help in other tasks such as scene completion. 

\subsection{Semantic Segmentation }
\subsubsection{Prologue and Significance}
This task relates to the labeling of each pixel in an image with its corresponding semantically meaningful category. Applications of semantic segmentation include domestic robots, content-based retrieval, self driving cars and medical imaging. Efforts to address the semantic segmentation problem have come a long way from  using hand crafted and data specific features to automatic feature learning techniques. Here, we summarize the important challenges for the problem and some of the most important methods for semantic segmentation that have had significant impact and inspired a great deal of research in this area. 

\subsubsection{Challenges}
Despite being an important task, segmentation is highly challenging because:
\begin{itemize}
\item Pixel level labeling requires both local and global information and challenge then is to design such algorithms that can incorporate the wide contextual information together.
\item The difficulty level increases a lot for the case of instance segmentation, where the same class is segmented into different instances.
\item Obtaining dense pixel level predictions, especially close to object boundaries, is challenging due to occlusions and confusing back-grounds. 
\item Segmentation is also affected by appearance, viewpoint and scale changes. 
\end{itemize}

\subsubsection{Methods Overview}
Traditionally, CRFs have been the default choice in the context of semantic segmentation \cite{cadena2013semantic, muller2014learning, hermans2014dense, deng2015semantic}. This is due to the reason that CRFs provide a flexible framework to model contextual information. As an example, \cite{hermans2014dense} exploit this property of CRF for the case of semantic segmentation of RGB-D images. They first developed a 2D semantic segmentation method based on decision forests \cite{breiman2001random} and  then transfered the 2D labels to 3D using a 3D CRF model to improve the RGB-D segmentation results. Other efforts to formulate semantic segmentation task into CRF framework include \cite{cadena2013semantic, muller2014learning,deng2015semantic}.
More recently, CNNs have been used to extract rich local features for 2.5D/3D image segmentation tasks \cite{couprie2014toward,Gupta2014, kalogerakis20163d, schneider2017multimodal}. A dominant trend in the deep learning based methods for semantic segmentation tasks has been to use  encoder-decoder networks in an end-to-end learnable pipeline, which enable high resolution segmentation maps \cite{long2015fully,badrinarayanan2015segnet,kendall2015bayesian,hazirbas2016fusenet}.

The work by Couprie et al. \cite{couprie2014toward} is among the pioneering efforts to use depth information along with RGB images for feature learning in semantic segmentation task. They used multi-scale convolutional neural networks (MCNN) for feature extraction that can be efficiently implemented using GPUs to operate in real-time during inference. Their work-flow involves fusing the RGB image with depth image using a Laplacian pyramid scheme which was then fed into the MCNN for feature extraction. The resulting features had a spatially low resolution, this was overcome using an up-sampling step. In parallel, RGB image was segmented into super-pixels. Final scene labeling was produced by aggregating the classifier predictions into the super-pixels. Note that although this approach was applied for video segmentation, it does not leverage temporal relationships and independently segments each frame. The real-time scene labeling of video sequences was achieved by using a computational efficient graph based scheme \cite{couprie2013causal} to compute temporal consistent super-pixels. This technique  was able to compute super-pixels in quasi-linear time, there by making it possible to use for real-time video segmentation.

Girshick et al. \cite{Girshick2014} presented a region CNN (R-CNN) method for detection and segmentation of RGB images which was later extended to RGB-D images \cite{Gupta2014}. The R-CNN \cite{Girshick2014} method extracts regions of interest from an input image, compute the features for each of the extracted regions using a CNN and then classify each region using class-specific linear SVMs. Gupta et al. \cite{Gupta2014} then extended the R-CNN method to RGB-D case by using a novel embedding for depth images. They purposed a geo-centric embedding called HHA to encode depth images using height above ground, horizontal disparity and angle with gravity for each pixel. They demonstrated that CNN can learn better features using the HHA embedding compared to raw depth images. Their proposed method \cite{Gupta2014} first uses multiscale combinatorial grouping (MCG) \cite{Arbelaez2014} to obtain region proposals from RGB-D images, followed by feature extraction using a CNN \cite{Alex2012} pre-trained on Imagenet \cite{imagenet2009} and fin-tuned on HHA encoded depth images. Finally, they pass these learned features of RGB and depth images through SVM classifier to perform object detection. They used superpixel classification framework \cite{gupta2013perceptual} on the output of the object detectors for semantic scene segmentation.

\begin{figure}[!t]
\centering
\includegraphics[width=\columnwidth]{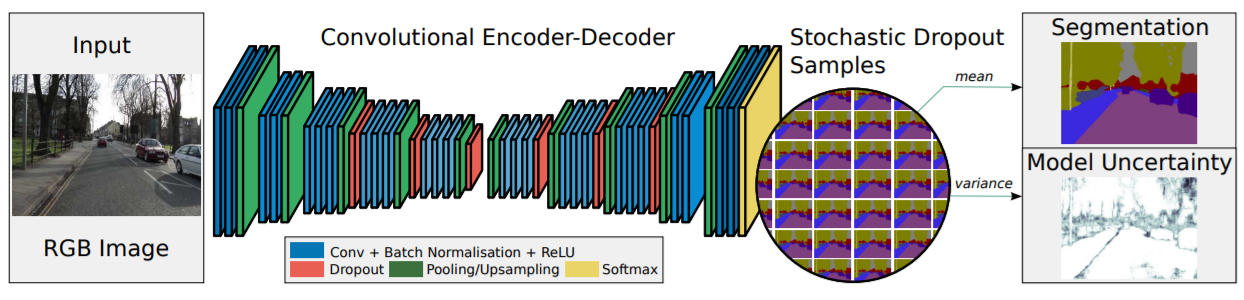}
\caption{A schematic of Bayesian encoder-decoder architecture for semantic segmentation with a measure of model uncertainty. (Courtesy of \cite{kendall2015bayesian}) }
\label{fig:b_segnet}
\end{figure}

Long et al. \cite{long2015fully} built an encoder-decoder architecture using a Fully Convolutional Network (FCN)  for pixel-wise semantic label prediction.  The network can take arbitrary sized input and produce corresponding sized outputs due to the fully convolutional architecture. \cite{long2015fully}  first redefined the pre-trained classification networks (AlexNet \cite{Alex2012}, VGG net \cite{simonyan2014very}, GoogLeNet \cite{szegedy2015going}) into their equivalent FCNs and therefore transferred their learned representation to the segmentation task. As one can expect, the FCNs based on classification nets downgrade the spatial resolution of visual information through consecutive sub-sampling operations. To improve the spatial resolution,  \cite{long2015fully} augments the FCN with a convolution transpose block for upsampling while keeping the end-to-end learning intact.  Further the final classification layer of each classification net \cite{Alex2012,simonyan2014very,szegedy2015going} was removed and the fully connected layers were replaced with 1x1 convolution followed by deconvolutional layer to upsample the output. To refine the predictions and for detailed segmentation, they introduced a skip architecture  which combined deep semantic information and shallow appearance information by  fusing the intermediate activations. To extend their method to RGB-D images, \cite{long2015fully} trained two networks: one for RGB images and a second for depth images represented by three dimensional HHA depth encoding introduced in \cite{Gupta2014}. The predictions from both nets are then summed at the final layer. After the successful application of FCNs \cite{long2015fully} to semantic segmentation, FCN based architectures since attracted a lot of attention from the research community and are extended to number of new tasks like region proposal \cite{ren2015faster}, contour detection \cite{xie2015holistically} and depth regression \cite{liu2016learning}. In a follow up paper \cite{shelhamer2017fully}, authors revisited the FCNs for semantic segmentation to further analyze, tune and improve results.

A measure of confidence based on which we can trust the semantic segmentation output of our model can be important in many important applications such as autonomous driving. None of the methods we discussed so far can produce probabilistic segmentation with a measure of model uncertainty. Kendall et al. \cite{kendall2015bayesian} came up with a framework to assign class labels to pixels with a measure of model uncertainty. Their method converts a convolutional encoder decoder network  \cite{badrinarayanan2015segnet} to Bayesian convolutional network that can produce probabilistic segmentation \cite{gal2016dropout} (see figure \ref{fig:b_segnet}). This technique can not only be used to convert many state of the art architecture like FCN \cite{long2015fully}, Segnet \cite{badrinarayanan2015segnet} and Dilation Network \cite{yu2015multi} to output probabilistic semantic segmentation but also improves the segmentation results by 2-3$\%$ \cite{kendall2015bayesian}. Their work \cite{kendall2015bayesian} is inspired by \cite{gal2015bayesian,gal2016dropout} where authors show that dropout \cite{srivastava2014dropout} can be used to approximate inference in a Bayesian neural network. \cite{gal2015bayesian} shows that dropout \cite{srivastava2014dropout} used at test time impose a Bernoulli distribution over the network’s filter weights by sampling the network with randomly dropped out units at the test time. This can be considered as obtaining Monte Carlo samples from the posterior distributions over the model. \cite{kendall2015bayesian} used this method to perform probabilistic inference over the segmentation model. It is important to note that softmax classifier produces relative probabilities between the class labels while the probability distribution from the Monte Carlo sampling \cite{kendall2015bayesian,gal2015bayesian,gal2016dropout} is an overall measure of the model’s uncertainty. 

\begin{figure}[!t]
\centering
\includegraphics[width=\linewidth]{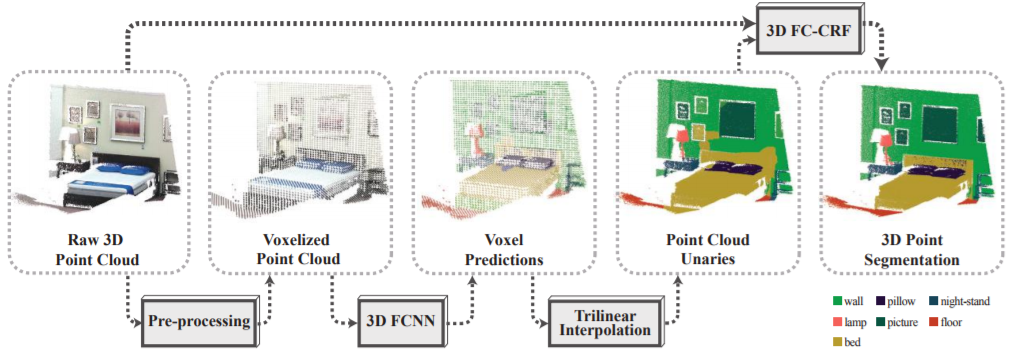}
\caption{SegCloud \cite{tchapmi2017segcloud} framework that takes a 3D point cloud as input, that is voxelized before feeding to 3D CNN. The voxelized representation is projected to the point cloud representation using trilinear interpolation. (Courtesy of \cite{tchapmi2017segcloud})}
\label{seg_cloud}
\end{figure}

Finally, we would like to mention that deep learning based models can learn to segment from irregular data representations e.g.,  consuming raw point clouds (with variable number of points) without the need of any voxelization or rendering. Qi et al. \cite{qi2016pointnet} developed a novel deep learning architecture called PointNet that can directly take point clouds as inputs and outputs segment labels for each point in the input. Subsequent works based on deep networks which directly operates on point clouds have also demonstrated excellent performance on the semantic segmentation task \cite{zeng20173dcontextnet,qi2017pointnet++}. Instead of solely using deep networks for context modeling, some recent efforts combine both CNN and CRFs for improved segmentations. As an example, a recent work on 3D point cloud segmentation combines the FCN and a fully connected CRF model which helps in better contextual modeling at each point in 3D \cite{tchapmi2017segcloud}. To enable a fully learnable system, the CRF is implemented as a differentiable recurrent network \cite{zheng2015conditional}. Local context is incorporate in the  proposed scheme by obtaining a voxelized representation at a coarse scale, and the predictions over voxels are used as the unary potentials in the CRF model (see Figure \ref{seg_cloud}).

We note that the encoder-decoder and dilation based architectures provide a natural solution to resolve the low resolution segmentation maps in RGBD based CNN architectures. Geometrically motivated encodings of raw depth information e.g., HHA encoding \cite{Gupta2014} can help improve model accuracy.
Finally, a measure of model uncertainty can be highly useful for practical applications which demand high safety. 

\subsection{Physics-based Reasoning}
\subsubsection{Prologue and Significance}
A scene is a static picture of the visual world. However, when humans look at the static image, they can infer hidden dynamics in a scene. As an example, from a still picture of a football field with players and a ball, we can understand the pre-existing motion patterns and guess the future events which are likely to happen in a scene. As a result, we can plan our moves and  take well-informed decisions. In line with this human cognitive ability, efforts have been made in computer vision to develop an insight into the underlying physical properties of a scene. These include estimating both current and future dynamics from a static scene \cite{mottaghi2016newtonian,wu2015galileo}, understanding the support relationships and stability of objects \cite{zheng2013beyond,silberman2012indoor,barbu2005generalizing}, volumetric and occlusion reasoning \cite{jia20153d,wang2013learning,bonde2014robust}. Applications of such algorithms include task and motion planning for robots, surveillance and monitoring.

\subsubsection{Challenges}
Key challenges for physics-based reasoning include:
\begin{itemize}
\item This task requires starting with very limited information (e.g., a still image) and performing extrapolation to predict rich information about scene dynamics.
\item A desirable characteristic is to adequately model prior information about the physical world. 
\item Physics based reasoning requires algorithms to reason about the contextual informations.
\end{itemize}

\subsubsection{Methods Overview}
\begin{figure*}[ht]
\centering
\includegraphics[width=0.9\linewidth, height=5cm]{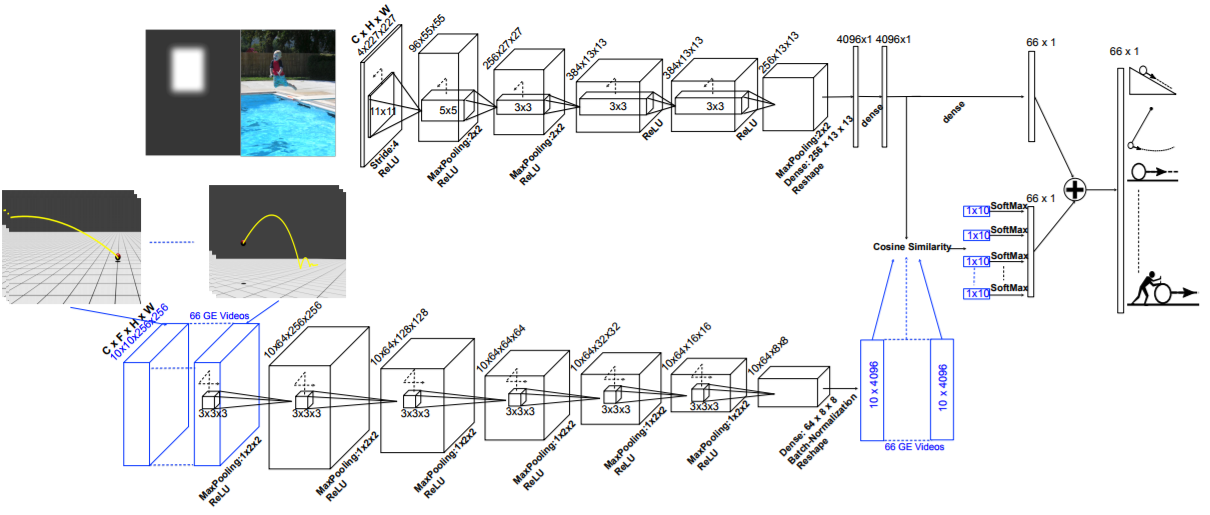}
\caption{Newtonian Neural Network ($N^3$) \cite{mottaghi2016newtonian}. The top stream of the architecture takes RGB image augmented with localization map of the targeted object and the bottom stream processes inputs from a game engine. Features from both streams are combined using cosine similarity and maximum response is used to find the scenario that best describes object motion in an image. (Courtesy of \cite{mottaghi2016newtonian})}
\label{N3}
\end{figure*}

\textbf{Dynamics Prediction:} Mottaghi et al. \cite{mottaghi2016newtonian} predicted the forces acting on an object and its future motion patterns to develop a deep physical understanding of a scene. To this end, they mapped a real world scenario to a set of 3D physical abstractions which model the motion of an object and the forces acting on it in the simplest terms e.g., a ball that is rolling, falling, bouncing or moving along a projectile.  This mapping is performed using a neural network with two branches, the first one processes a 2D real image while the other one processes 3d abstractions. The 3D abstractions were obtained from game rendering engines and their corresponding RGB, depth, surface normal and optical flow data was fed to the deep network as input. Based on the mapped 3D abstraction, long term motion patterns from a static image were predicted (see Figure \ref{N3}).

 Wu et al. \cite{wu2015galileo} proposed a generative model  based on  behavioral studies with the argument that physical scene understanding developed by human brain is a simulation of a mental physics engine \cite{battaglia2013simulation}. The mental engine carries physical information about the world objects and the Newtonian laws they obey, and performs simulations to understand and infer scene dynamics. The proposed generative model, called `\emph{Galileo}', predicts physical attributes of objects (i.e., 3D shape  position, mass and friction) by taking the feedback from a physics engine which estimates the future scene dynamics by performing simulations. An interesting aspect of this work is that a deep network was trained using the predictions from the Galileo, resulting in a model which can efficiently predict physical properties of objects and future scene dynamics in static images.

\textbf{Support Relationships:} Alongside the hidden dynamics in a scene, there exist rich physical relationships in a scene which are important for scene understanding. As an example, a book on a table will be supported by the table surface and the table will be supported by the floor or the wall. These support relationships are important for robotic manipulation and interaction in man-made environments.  Silberman et al. \cite{silberman2012indoor} proposed a CRF model to segment cluttered indoor environments and identify support relationships between the objects in RGB-D imagery. They categorized a scene into four geometric classes, namely \emph{ground}, \emph{fixed structures} (e.g., walls, ceiling), \emph{furniture} (e.g., cabinet, tables) and props (small moveable objects). The overall energy function incorporated both local features (e.g., appearance cues) and pairwise interactions between objects (e.g., physical proximity). An integer programming formulation was introduced to efficiently minimize the energy function. The incorporation of support relationship information has been shown to improve the performance on other relevant tasks such as the scene parsing \cite{jia20133d,zhuo2017indoor}.  

\begin{figure}[!h]
\centering
\includegraphics[width=\linewidth]{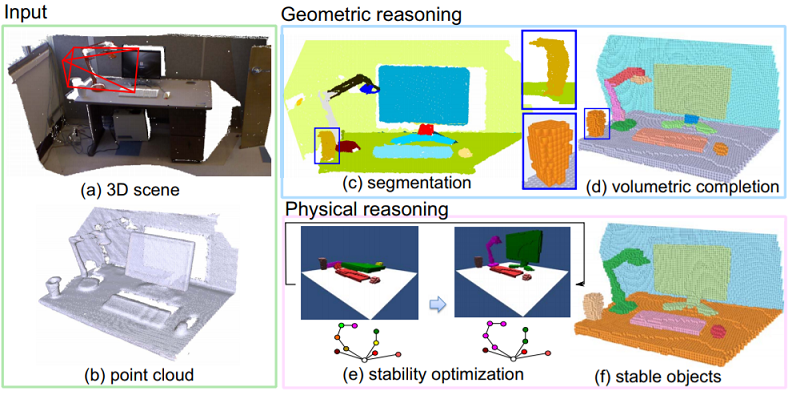}
\caption{A 3D scene converted into point cloud before feeding to geometric and physical reasoning module \cite{zheng2013beyond}. (Courtesy of \cite{zheng2013beyond})}
\label{beyond_point_cloud}
\end{figure}

\textbf{Stability Analysis:} In static scenes, it is highly unlikely to find objects which are unstable with respect to gravity. This physical concept has been employed in scene understanding to recover geometrically and physically stable objects and scene parses. Note that the support relationships predicted in \cite{silberman2012indoor} does not ensure the physical stability of objects. 
Zheng et al. \cite{zheng2013beyond} reasoned about the stability of 3D volumetric shapes, which were recovered from a either a sparse or a dense 3D point cloud of indoor scenes (available from range sensors). A parse graph was built such that each primitive was constrained to be stable under gravity, and the falling primitives were grouped together to form stable candidates. The graph labeling problem was solved using the Swendsen-Wang Cut partitioning algorithm \cite{barbu2005generalizing}.  They noted that such a reasoning helps in achieving better performance on linked tasks such as object segmentation and 3D volume completion (see Figure \ref{beyond_point_cloud}).

While \cite{zheng2013beyond} performed physical reasoning utilizing mainly depth information,  Jia et al. \cite{jia20153d}  incorporate both color and depth data for such analysis. Similar to  \cite{zheng2013beyond}, \cite{jia20153d} also fits 3D volumetric shapes on RGB-D images and performs physics-based reasoning by considering their 3D intersections, support relationships and stability.  As an example, a plausible explanation of a scene is the one, where 3D shapes cannot overlap each other, are supported by each other and will not fall under gravity. An energy function was defined over the over-segmented image and a number of unary and pairwise features were used to account for stability, support, appearance and volumetric characteristics. The energy function was minimized using a randomized sampling approach \cite{chang2011efficient} which either splits or merges the individual segments to obtain improved segmentations. They used the physical information for semantic scene segmentation, where it was shown to improve the performance.

\textbf{Hazard Detection:} An interesting direction from the previous works is to predict which objects can potentially fall in a scene. This can be highly useful to ensure safety and avoid accidents in work places (e.g., a construction site), domestic environments (e.g., child care) and due to natural disasters (e.g., earth quake). Zheng et al. \cite{zheng2014detecting} first estimated potential causes of disturbance (i.e. human activity and natural disasters) and then predicted the potentially unstable objects which can fall as a result of disturbance. Given a 3D point cloud, a `\emph{disturbance field}' is predicted for a possible type of disturbance (e.g. using motion capture data for human movement) and its effect is estimated using the mechanics principles (e.g., conservation of energy and momentum after collision). In terms of predicting scene dynamics, this approach goes beyond inferring motions from the given static image, rather it considers ``what if?" scenarios and predicts associated dynamics. More recently, Dupre et al.\cite{dupre2017automated} have proposed to use CNN  to perform automatic risk assessment in scenes.

\textbf{Occlusion Reasoning:}
Occlusion relationship is another important physical and contextual cue, that commonly appears in cluttered scenes. Wang et al. \cite{wang2013learning} showed that occlusion reasoning helps in object detection. They introduced a hough voting scheme, which uses depth context at multiple levels (e.g., object relationship with near-by, far-away and occlusion patches) in the model to jointly predict the object centroid and its visibility mask. They used a dictionary learning approach based on local features such as Histogram of Gradients (HOG) \cite{felzenszwalb2008discriminatively} and Textons \cite{shotton2006textonboost}. An interesting result was that the occlusion relationships are important contextual cues which can be useful for object detection and segmentation. In another subsequent work, Bonde et al. \cite{bonde2014robust} used occlusion information computed from depth data to recognize individual object instances. They used a random decision forest classifier trained using a max-margin objective to improve the recognition performance.

\subsection{Object Pose Estimation}
\subsubsection{Prologue and Significance}
The pose estimation task deals with finding object's position and orientation with respect to a specific coordinate system. Information about an object's pose is crucial for object manipulation by robotic platforms and scene reconstruction e.g., by fitting 3D CAD models. Note that the pose estimation task is highly related to the object detection task, therefore existing works address both problems sequentially \cite{schwarz2015rgb} or in a joint framework \cite{mottaghi2015coarse,sedaghat2016orientation,brachmann2014learning}. Direct feature matching techniques (e.g., between images and models) have also been explored for pose estimation \cite{lim2013parsing,tejani2014latent}.

\subsubsection{Challenges}
Important difficulties that pose estimation algorithms encounter are:
\begin{itemize}
\item The requirement of detecting objects and estimating their orientation at the same makes this task particularly challenging.
\item Object's pose can vary significantly from one scene to another, therefore algorithm should be invariant to these changes.
\item Occlusions and deformations make the pose estimation task difficult especially when multiple objects are simultaneously present. 
\end{itemize}

\subsubsection{Methods Overview}
Lim et al. \cite{lim2013parsing} used 3D object models to estimate object pose in a given image. Object appearances can change from one scene to another due to number of factors including geometric deformation and occlusions. The challenge then is not only to retrieve the relevant model for an object but to accurately fit it to real images. Their proposed algorithm takes key detectors like geometric distances, their local correspondence and global alignment to find candidate poses. Tejani et al. \cite{tejani2014latent} proposed a method for 3D object detection and pose estimation which is robust to foreground occlusion and background clutter. The presented framework called Latent-Class Hough Forests (LCHF) is based on a patch based detector called Hough Forests \cite{gall2011hough} and trained on only positive data samples of 3D synthetic model renderings. They used LINEMOD \cite{hinterstoisser2011multimodal}, a 3D holistic template descriptor, for patch representation and integrate it into random forest framework using template-based splitting function. At test time, class distributions are iteratively inferred to jointly estimate 3D object detection, pose and pixel-wise visibility map.

\begin{figure}[!htp]
\centering
\includegraphics[width=\linewidth, height=3cm]{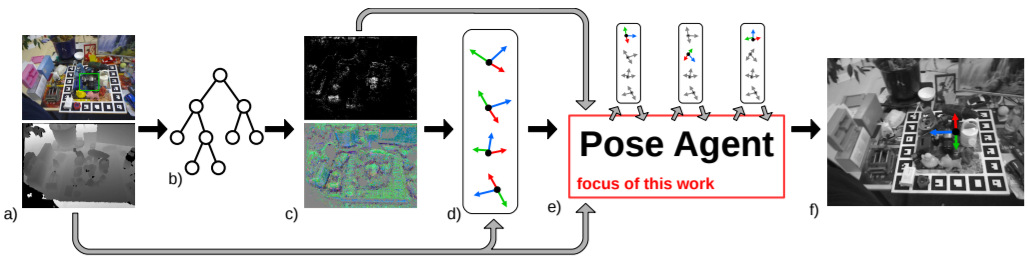}
\caption{Pose estimation pipeline as proposed by \cite{krull2016poseagent}. RGB-D input is first processed by random forest. These predictions are used to find pose candidates then a reinforcement agent refines these candidates to find the best pose. (Courtesy of \cite{krull2016poseagent})}
\label{pose_reinforcement}
\end{figure}

\begin{figure*}[htp]
\centering
\includegraphics[width=0.85\linewidth]{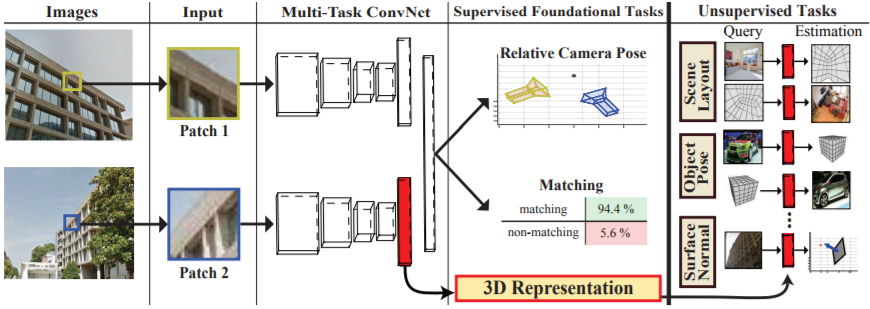}
\caption{CNN trained to learn generic features by matching image patches with their corresponding camera angle. These generic feature representation can be used for multiple tasks at test time including object pose estimation. (Courtesy of \cite{zamir2016generic})}
\label{pose_generic}
\end{figure*}

Mottaghi et al. \cite{mottaghi2015coarse} argued that object detection, 3D pose estimation and sub-category recognition are correlated tasks.  One task can provide complimentary information to better understand the others, therefore, they  introduced a hierarchal method based on a hybrid random field model that can handle both continuous and discrete variables to jointly tackle these three tasks. The main idea here is to represent objects in a hierarchal fashion such that the top-layer captures high level coarse information e.g., discrete view point and object rough location and layers below capture more accurate and refined information e.g., continuous view point and object category (e.g., a car) and sub-category (e.g., specific type of a car). Similar to \cite{mottaghi2015coarse}, Brannchman et al. \cite{brachmann2014learning} proposed a method to jointly estimate object class and its 6D pose (3D rotation, 3D translation) in a given RGB-D image. They trained a decision forest with 20 objects under two different lighting conditions and a set of background images. The forest consists of three trees that use color and depth difference features to jointly learn 3D object coordinates and object instances probabilities. A distinguishing feature of their approach is the ability to scale to both textured and texture-less objects.

The basic idea behind methods like \cite{brachmann2014learning} is to generate number of interpretations or sample pose hypothesis and then find the one that best describe an object pose. \cite{brachmann2014learning} achieves this by minimizing an energy function using RANSAC. \cite{krull2015learning} built upon the idea in \cite{brachmann2014learning}, but used a CNN trained with a probabilistic approach to find the best pose hypothesis. \cite{krull2015learning} generated pose hypothesis, scored it based on their quality and decided which hypothesis to explore next. This sort of decision making is non-differentiable and does not allow an end-to-end learning framework. Krull et al. \cite{krull2016poseagent} improved upon the work presented in \cite{krull2015learning} by introducing a reinforcement learning to incorporate non-differentiable decision making into an end-to-end learning framework (see Figure \ref{pose_reinforcement}).

 To benefit from representation power of CNN, Schwarz et al. \cite{schwarz2015rgb} used AlexNet\cite{Alex2012}, a large-scale CNN model trained on ImageNet visual recognition dataset. This model was used to extract features for object detection and pose estimation in RGB-D images. The novelty in their work is the pre-processing of color and depth images. Their algorithm segments objects form a given RGB image and removes the background. They colorized the depth image based on the distance from the object center. Both processed images are then fed to the AlexNet to extract features which are then concatenated before passing through a SVM classifier for object detection and a support vector regressor (SVR) for pose estimation. 
 
 One major challenge for pose estimation is that it can vary significantly from one image to another. To address this issue, Wohlhart et al. \cite{wohlhart2015learning} proposed to create clusters of CNN based features, that are indicative of object category and their poses. The idea is to generate multiple views of each object in the database, then each object view is represented by a learned descriptor that stores information about object identity and its pose. The CNN is trained under euclidean distance constraints such that the distance between descriptors of different objects is large and the distance between descriptors of same object is small but still provides a measure of differences in pose. In this manner, clusters of object labels and poses are formed in the descriptor space. At test time, a nearest neighbor search was used to find similar descriptor for a given object. To further improve and tackle the the pose variation issue in an end-to-end fashion,  \cite{doumanoglou2016siamese} formulated the pose estimation problem as regression task and introduced an end-to-end Siamese learning framework. Angles variations of the same object in multiple images are tackled by using Siamese network architecture with novel loss function to enforce similarity between the features of given training images and their corresponding poses.

CNN models have an extraordinary ability to learn generic representations that are transferable across tasks. Zamir et al. \cite{zamir2016generic} validated this by training a CNN to learn 3D generic representations to simultaneously address multiple tasks. In this regard, they trained a multi-task CNN to jointly learn camera pose estimation and key point matching across extreme poses. They showed with extensive experimentation that internal representation of such a trained CNN can be used for other predictions tasks such as object pose, scene layout and surface normal estimation (see Figure \ref{pose_generic}).
Another important approach on pose estimation was introduced in \cite{kehl2016deep}, where instead of full images, a network was trained on image patches. Kehl et al.
\cite{kehl2016deep} proposed a framework that consists of sampling the scene at discrete steps and extracting local and scale invariant RGB-D patches. For each patch, they compute its deep-regressed feature using a trained auto-encoder network and perform K-NN search with a codebook of local object patches. Each codebook entry holds a local 6D vote and is cast into Hough space only to survive a confident threshold. The codebook entries are coming from densely sampled synthetic views. Each entry stores its deep-regression feature and a 6D local vote. They employ a convolutional auto-encoder (CAE) that has been trained on 1.5M local RGB-D patches.

\begin{figure*}[htp]
\centering
\includegraphics[width=0.8\linewidth]{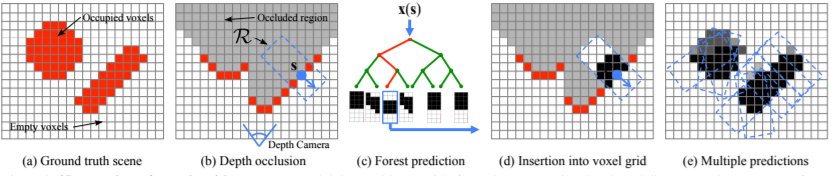}
\caption{A Random forest based framework to predict 3D geometry \cite{firman2016structured}. (Courtesy of \cite{firman2016structured})}
\label{firman_random}
\end{figure*}

\begin{figure*}[htp]
\centering
\includegraphics[width=\linewidth]{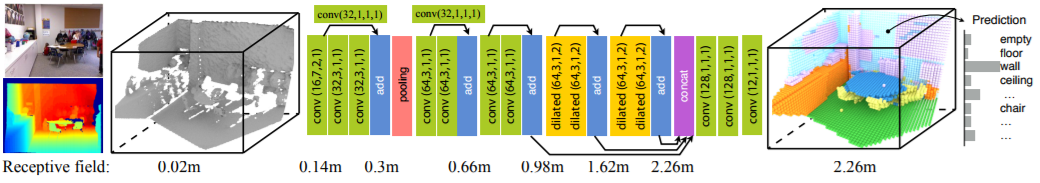}
\caption{SSCNet architecture \cite{song2016semantic} trained to reconstruct complete 3D scene from a single depth image. (Courtesy of \cite{song2016semantic})}
\label{SSCNet}
\end{figure*}
\subsection{3D Reconstruction from RGB-D}
\subsubsection{Prologue and Significance}
Humans visualize and interpret surrounding environments in 3D. The 3D reasoning about an object or a scene allows a deeper understanding of the mechanics, shape and 3D texture characteristics. For this purpose, it is often desirable to recover the full 3D shape from a single or multiple RGB-D images. 3D reconstruction is useful in many applications areas including medical imaging, virtual reality and computer graphics. Since, the 3D reconstruction from densely overlapping RGB-D views of an object \cite{whelan2012kintinuous,newcombe2011kinectfusion} is relatively a simpler problem, here we focus on scene reconstruction from either a single or a set of RGB-D images with partial occlusions leading to incomplete information.  .

\subsubsection{Challenges}
3D reconstruction is highly challenging problem because:
\begin{itemize}
\item Complete 3D reconstruction from incomplete information is an ill-posed problem with no unique solution. 
\item This problem poses a significant challenge due to sensor noise, low depth resolution, missing data and quantization errors.
\item It requires appropriately incorporating external information about the scene or object geometry for a successful reconstruction.
\end{itemize}

\subsubsection{Methods Overview}
3D reconstruction from a single RGB-D image has recently gained popularity due to the availability of cheap depth sensors and powerful representation learning networks. This task is also called as the `\emph{shape or volumetric completion}' task, since a RGB-D image provides a sparse and incomplete point cloud which is completed to produce a full 3D output. For this task, CRF models have been a natural choice because of their flexibility to encode geometric and stability relationships to generate physically viable outputs \cite{kim20133d,zheng2013beyond}. Specifically, Kim et al. \cite{kim20133d} proposed a CRF model defined over voxels to jointly reconstruct 3D volumetric output along with the semantic category labels for each voxel. Such a joint formulation helps in modeling the complex interplay between semantic and geometric information in a scene. Firman et al. \cite{firman2016structured} proposed a structured prediction framework developed using a Random Forest to predict the 3D geometry given the observed incomplete shapes. Unlike \cite{kim20133d}, a shortcoming of their model is that it does not uses semantic details of voxels alongside the geometric information (see Figure \ref{firman_random}).

With the success of deep learning, the above mentioned ideas have recently been formulated as end-to-end trainable networks with several interesting extensions. As an example, Song et al. \cite{song2016semantic} proposed a 3D CNN to jointly perform semantic voxel labeling and scene completion from a single RGB-D image. The CNN architecture makes use of successful ideas in deep learning such as skip connections \cite{he2016deep} and dilated convolutions \cite{yu2015multi} to aggregate scene context and the use of a large-scale dataset (SUNCG) (see Figure \ref{SSCNet}). A convolutional LSTM based recurrent network has been proposed in \cite{choy20163d} for 3D reconstruction  of individual objects (in contrast to complete scenes as in \cite{song2016semantic}). First, an object view is encoded followed by learning a representation using LSTM which is then used for decoding. The benefit of this approach is that the latent representations can be stored in the memory (LSTM) and updated if more views of an object become available. Another similar approach for shape completion uses first an encoder-decoder architecture to obtain a coarse 3D output which is refined using  similar high-resolution shapes available as prior knowledge \cite{dai2016shape}. This incorporates both bottom-up and top-down  knowledge transfer (i.e. using shape category information along with the incomplete input) to recover better quality 3D outputs. Gupta et al. \cite{gupta2015aligning} investigated a similar data-driven approach by first identifying individual object instances in a scene and using a library of common indoor objects to retrieve and align the 3D model with the given RGB-D object instance. This approach, however, does not implement an end-to-end learnable pipeline and focuses only on object reconstruction instead of a full scene reconstruction.

Early work on 3D reconstruction from multiple overlapping RGB-D images used the concept of averaging TSDF obtained from each of the RGB-D views \cite{hoppe1996progressive, choi2015robust}. However, the TSDF based reconstruction techniques require a large number of highly overlapping images due to their inability to complete occluded shapes. More recently, OctNet based representations have been used in \cite{riegler2017octnetfusion}, which allow scaling 3D CNNs to work on considerably high dimensional volumetric inputs compared to the regular voxel based models \cite{wu2016single}. The octree based representations take account of empty spaces in 3D environments and use low spatial resolution voxels for the unoccupied regions, thus leading to faster processing with high resolutions in 3D deep networks \cite{Riegler2017OctNet}. In contrast to the regular octree based methods \cite{Riegler2017OctNet}, the scene completion task requires the prediction of reconstructed scene along with a suitable 3D partitioning for the octree representation. These two outputs are predicted using a u-shaped 3D encoder-decoder network in \cite{riegler2017octnetfusion}, where the short-cut connections exist between the corresponding coarse-to-fine layers in the encoder and the decoder modules.

\subsection{Saliency Prediction}
\subsubsection{Prologue and Significance}
The human visual system selectively attends to salient parts of a scene and performs a detailed understanding for the most salient regions. The detection of salient regions corresponds to important objects and events in a scene and their mutual relationships. In this section, we will review saliency estimation approaches which use a variety of 2.5/3D sensing modalities including RGB-D \cite{lang2012depth,peng2014rgbd}, stereopsis \cite{niu2012leveraging,fang2014saliency}, light-field imaging \cite{li2014saliency} and point-clouds \cite{kobyshev20163d}. Saliency prediction is valuable in several applications e.g., user experience analysis, scene summarization, automatic image/video tagging, preferential processing on resource constrained devices, object tracking and novelty detection.

\subsubsection{Challenges}
Important problems for the saliency prediction task are:
\begin{itemize}
\item Saliency is a complex function of different factors including appearance, texture, background properties, location, depth etc. It is a challenge to model these intricate relationships. 
\item It requires both top-down and bottom-up cues to accurately model objects saliency. 
\item An key requisite is to adequately encode the local and global context. 
\end{itemize}

\subsubsection{Methods Overview}
Lang et al. \cite{lang2012depth} were the first to introduce a RGB-D and 3D dataset (NUS-3DSaliency) with corresponding eye-fixation data from human viewers. They analyzed the differences between the human attention maps for 2D and 3D data and found depth to be an important cue for visual attention (e.g., attention is focused more on nearby depth ranges). To learn the relationships between visual saliency and depth, a generative model was trained to learn their joint distributions. They showed that the incorporation of depth resulted in a consistent improvement for previous saliency detection methods designed for 2D images. 

\begin{figure*}[!htp]
\centering
\includegraphics[width=\linewidth, height=6cm]{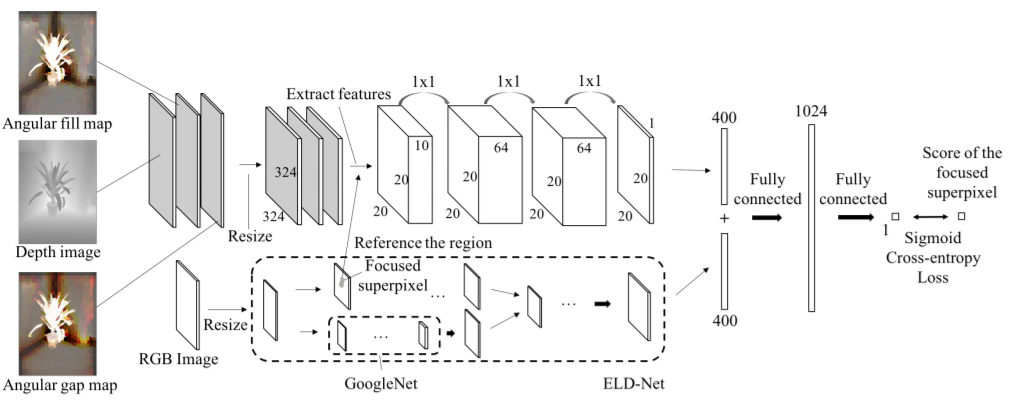}
\caption{Network architecture for saliency prediction \cite{shigematsu2017learning}. RGB saliency features are fused with super-pixel based handcrafted features to get the overall saliency score. (Courtesy of \cite{shigematsu2017learning})}
\label{ELD-Net}
\end{figure*}

Based on the insight that salient objects  are likely to appear at different depths, Peng et al. \cite{peng2014rgbd} proposed a multi-stage model where local, global and background contrast-based  cues  were used to predict a rough estimate of saliency. This initial saliency estimate was used to calculate a foreground probability map which was combined with an object prior to generate final saliency predictions. The proposed method was evaluated on a newly-introduced large-scale benchmark dataset.
In another similar approach, Coptadi et al. \cite{ciptadi2013depth} calculated local 3D shape and layout features (e.g., plane and normal cues) using the depth information to improve object saliency. Feng et al. \cite{feng2016local} advocated that simple depth-contrast based features create confusions for rich backgrounds. They  proposed a new descriptor which measures the enclosure provided by the background to fore-ground salient objects.

More recently, \cite{qu2017rgbd} proposed to use a CNN for RGB-D saliency prediction. However, their approach is not end-to-end trainable as they first extract several hand-crafted features and fuse them together followed by an off-line smoothing and saliency prediction stage with in a CNN. Shigematsu et al. \cite{shigematsu2017learning} extended a RGB based saliency detection network (ELD-Net \cite{lee2016deep}) for the case of RGBD saliency detection (see Figure \ref{ELD-Net}). They augmented the high-level feature description from a pre-trained CNN with a number of low-level feature descriptions such as the depth contrast, angular disparity and background enclosure \cite{feng2016local}. Due to the limited size of available RGB-D datasets, this technique relies on the weights learned on color image based saliency datasets. 

Stereoscopic images provide approximate depth information based on the disparity map between a pair of images. This additional information has been shown to assist in visual saliency detection especially when the salient objects do not carry significant color and texture cues.  Niu et al. \cite{niu2012leveraging} introduced an approach based on the disparity contrast to incorporate depth information in saliency detection. Further, a prior was introduced based on domain knowledge in stereoscopic photography, which prefers regions inside the viewing comfort zone to be more salient. Fang et al. \cite{fang2014saliency} developed on similar lines and used appearance as well as depth feature contrast from stereo images. A Gaussian model was used to weight the distance between the patches, such that both the global and local contrast can be accounted for saliency detection.

Plenoptic camera technology can capture the light field of a scene, which provides both the intensity and direction of the light rays. Light field cameras provide the flexibility to refocus after photo capture and can provide a depth map in both indoor and outdoor environments. Li et al. \cite{li2014saliency} used the focusness and depth information available from light field cameras to improve saliency detection. Specifically, frequency domain analysis was performed to measure focusness in each image. This information was used alongside depth to estimate foreground and background regions, which were subsequently improved using contrast and objectness measures.

In contrast to the above techniques which mainly augment depth information for saliency prediction, \cite{kobyshev20163d} detected salient patterns in 3D city-scale point clouds to identify land-mark buildings. To this end, they introduced a distance measure which quantifies the uniqueness of a landmark by considering its distinctiveness compared to the neighborhood. While this work is focused on outdoor man-made structures, to the best of our knowledge, the problem of finding salient objects in indoor 3D scans is not investigated in the literature.

\subsection{Affordance Prediction}
\subsubsection{Prologue and Significance}
Object based relationships (e.g., chairs are close to desk) have been used with success in scene understanding tasks such as semantic segmentation and holistic reasoning \cite{zhang2016deepcontext}. However, an interesting direction to interpret indoor scenes is by understanding the functionality or affordances of objects \cite{gibson2014ecological} i.e., what actions can be performed on a particular object  (e.g., one can sit on a chair, place a coffee cup on a table). These characteristics of objects can be used as attributes, which have been found to be useful to transfer knowledge across categories \cite{farhadi2009describing}. Such a capability is important in application domains such as assistive, domestic and industrial  robotics, where the robots need to actively interact with the surrounding environments. 

\subsubsection{Challenges}
Affordance detection is challenging because:
\begin{itemize}
\item  This task requires information from multiple sources and reasons about the content to discover relationships.
\item It often requires modeling the hidden context (e.g., humans not present in the scene) to predict the correct affordances of objects. 
\item Reasoning about physical and material properties is crucial for this affordance detection. 
\end{itemize}

\subsubsection{Methods overview}
A seminal work on affordance reasoning by Grabner et al. \cite{grabner2011makes} estimated places where a person can `sit' in an indoor 3D scene. Their key idea was to predict affordance attributes by assuming the presence of an interacting entity i.e., a human. The functional attributes proved to be a complementary source of information which in turn improved the 'chair' detection performance. On similar lines, Jiang et al. \cite{jiang2013hallucinated,jiang2016modeling} hallucinated humans in indoor environments to predict the human-object relationships. To this end, a latent CRF model was introduced, which jointly infers the human pose and object affordances. The proposed  probabilistic graphical model was composed of objects as nodes and their relationships were encoded as graph edges. Alongside these, latent variables were used to represent hidden human context. The relationships between object and humans were used to perform 3D semantic labeling of point clouds. 

Koppula and Saxena \cite{koppula2016anticipating} used object affordances in a RGB-D image based CRF model to forecast the future human actions so that an assistive robot can generate a response in time. \cite{pieropan2015functional} suggested affordance descriptors which model the way an object is operated by a human in a RGB-D video. The above mentioned approaches deal with the affordance prediction for generic objects. Myers et al. \cite{myers2015affordance} introduced a new dataset comprising of everyday use tools (e.g., hammer, knife). A given image was first divided into super-pixels, followed by computation of a number of geometric features such as normals and curvedness. A sparse coding based dictionary learning approach was used to identify parts and predict the corresponding affordances. For large dictionary sizes, such an approach can be quite computationally expensive, therefore a random forest based classifier was proposed for real-time applications. Note that all of the approaches mentioned so far, including \cite{myers2015affordance}, used hand-crafted features for affordance prediction. 

\begin{figure}[htp]
\centering
\includegraphics[width=\linewidth]{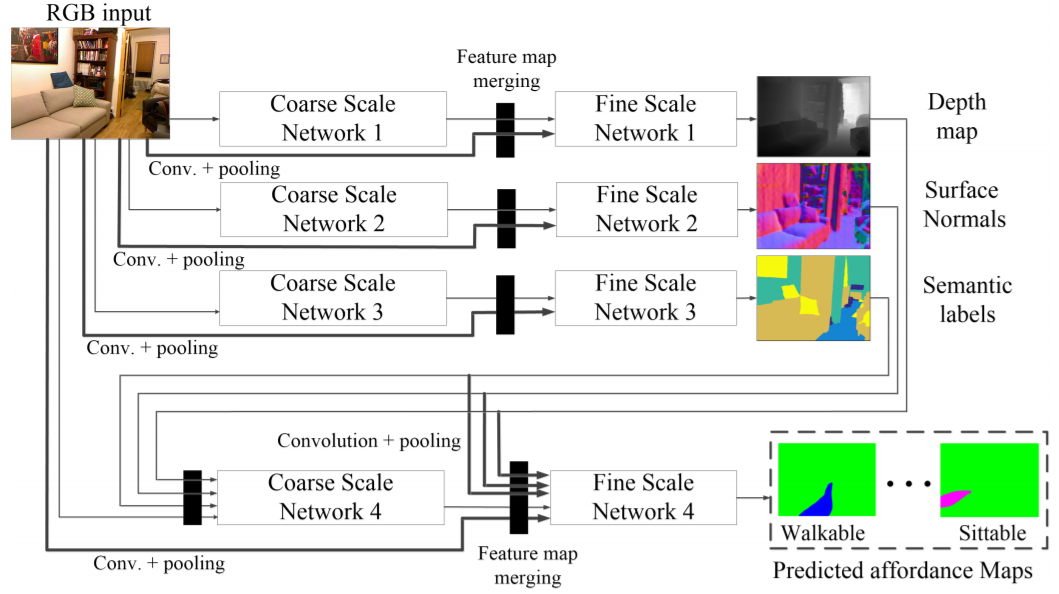}
\caption{A multi-scale CNN proposed in \cite{roy2016multi}. The coarse scale network extracts global representations encoding wide context while the fine-scale network extracts local representations such as object boundaries. Affordance labels are predicted by combining both representations. (Courtesy of \cite{roy2016multi})}
\label{afford_multi_cnn}
\end{figure}

More recently, automatic feature learning mechanisms such as CNNs have been used for object affordance prediction \cite{ye2017can,roy2016multi,nguyen2016detecting}. Nquyen et al. \cite{nguyen2016detecting} proposed a convolutional encoder-decoder architecture to predict grasp affordances for tools using RGB-D images. The network input was encoded as a HHA encoding \cite{Gupta2014} of depth along with the color image. A more generic affordance prediction framework was presented in \cite{roy2016multi}, which used  a multi-scale CNN to provide affordance segmentations for indoor scenes (see Figure \ref{afford_multi_cnn}). Their architecture explicitly used mid-level geometric and semantic representations such as labels, surface normals and depth maps at coarse and fine levels to effectively aggregate information. Ye et al. \cite{ye2017can} framed the affordance prediction problem as a region detection task and used the VGGnet CNN to categorize a region into a rich set of functional classes e.g., whether a region can afford an open, move, sit or manipulate operation. 

\begin{figure*}[htp]
\centering
\includegraphics[width=\linewidth]{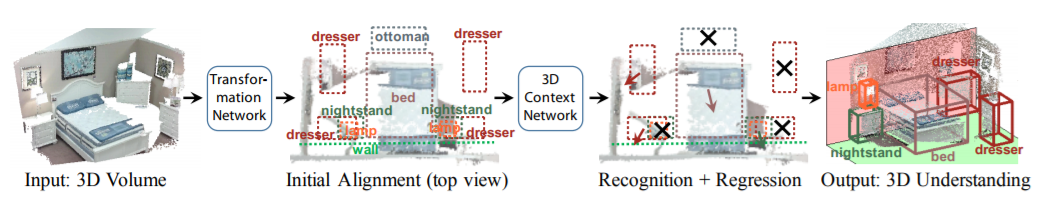}
\caption{3D scene understanding framework as proposed by \cite{zhang2016deepcontext}. A volumetric representation is first derived from depth image and aligned with the input data, next a 3D CNN estimates the objects presence and adjust them based on holistic scene features for full 3D scene understanding (Courtesy of \cite{zhang2016deepcontext}).}
\label{holistic_deep}
\end{figure*}

\begin{figure}[htp]
\centering
\includegraphics[width=\linewidth]{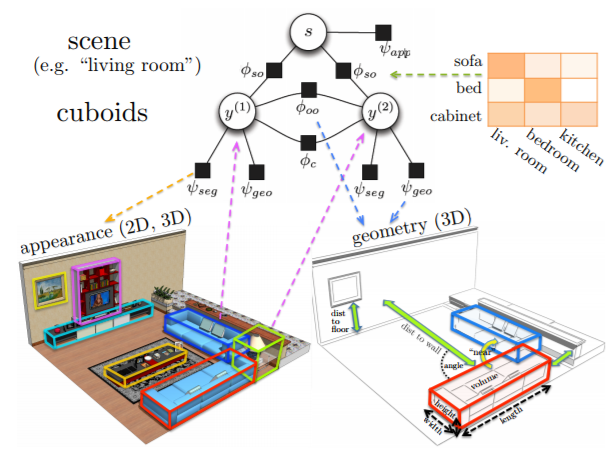}
\caption{Appearance and geometric properties are represented using object cuboids which are then used to define scene to object and object to object relations. This information is integrated by a CRF model for holistic scene understanding. (Courtesy of \cite{lin2013holistic}}
\label{holistic_rgbd_cameras}
\end{figure}

\subsection{Holistic/Hybrid Approaches}
\subsubsection{Prologue and Significance}
Up till now, we have covered individual tasks that are important to develop an understanding about e.g., scene semantics, constituent objects and their locations, object functionalities and their saliency.  In holistic scene understanding, a model aims to simultaneously reason about multiple complimentary aspects of a scene to provide a detailed scene understanding. Such an integration of individual tasks can lead to practical systems which require joint reasoning such as robotic platforms interacting with the real world (e.g., automated systems for hazard detection and quick response and rescue). In this section, we will review some of the significant efforts for holistic 2.5D/3D scene understanding. We will outline the important challenges and explore different ways the information from multiple sources is integrated in the literature for the specific case of indoor scenes \cite{li2010towards,lin2013holistic,choi2013understanding,zhuo2017indoor,gupta2015indoor}. 

\subsubsection{Challenges}
Important obstacles for holistic scene understanding are:
\begin{itemize}
\item Accurately modeling relationships between objects and background is a hard task in real-world environments due to the complexity of inter-object interactions. 
\item Efficient training and inference is difficult due to the requirement of reasoning at multiple levels of scene decomposition.
\item Integration of multiple individual tasks and complementing one source of information with another is a key challenge. 
\end{itemize}

\subsubsection{Methods Overview}
Li et al. \cite{li2010towards} proposed a Feedback Enabled Cascaded Classification Model (FE-CCM), which combines individual classifiers trained for a specific task e.g., object, event detection, scene classification and saliency prediction.  This combination is performed in a cascaded fashion with a feedback mechanism to jointly learn all task specific models for scene understanding and robot grasping. They argued that with the feedback mechanism, FE-CCM learns meaningful relationships between sub-tasks. An important benefit of FE-CCM \cite{li2010towards} is that it can be trained on heterogeneous datasets meaning it does not require data points to have labels for all the tasks. Similar to \cite{li2010towards}, a two-layer generic model with a feedback mechanism was presented in \cite{li2010generic}. In contrast to the above methods, \cite{lin2013holistic} presented a holistic graphical model (a CRF) to integrate scene geometry, relations between objects, interaction of objects with scene environment for 3D object recognition (see Figure \ref{holistic_rgbd_cameras}). They extended Constrained Parametric Min-Cuts (CPMC) \cite{carreira2012cpmc} method to generate cuboids from RGB-D images. These cuboids contain information about scene geometry, appearance and help in modeling contextual information for objects.

To understand complex scenes, it is desirable to learn interactions between scene elements e.g., scene structures-to-object interaction and object-to object interaction. Choi et al. \cite{choi2013understanding} proposed a method that can learn these scene interactions and integrate information at multiple levels to estimate scene composition. Their hierarchal scene model learns to reason about complex scenes by fusing together scene classification, layout estimation and object detection tasks. The model takes a single image as an input and generates a parse graph that best fit the image observations. The graph root represents the scene category and layout, while the graph leaves represent objects detections. In between, they introduced novel 3D Geometric Phrases (3DGP) that encode semantic and geometric relations between objects. A Reversible Jump Markov Chain Monte Carlo (RJMCMC) sampling technique was used to search for the best fit graph for the given image.

The wide contextual information available from human eyes plays a critical role in human scene understanding. Field of view (FOV) of a typical camera is only 15$\%$ compared to human vision system.  Zhang et al. \cite{zhang2014panocontext} argued that due to limited FOV, a typical camera cannot capture full details presented in a scene e.g., number of objects or occurrences of an object. Therefore, a model built on single images with limited FOV cannot exploit full contextual information in a scene. Their proposed method takes a 360-degree panorama view and generates 3D box representation of the room layout along with all of the major objects. Multiple 3D representations are generated using variety of image characteristics and then a SVM classifier is used to find the best one.

Zhang et. al \cite{zhang2016deepcontext} developed a 3D deep learning architecture to jointly learn furniture category and location from a single depth image (see Figure \ref{holistic_deep}). They introduced a template representation for 3D scene to be consumed by a deep net for learning. A scene template encodes a set of objects and their contextual information in the scene. After training, their so called DeepContext net learns to recognize multiple objects and their locations based on both local object and contextual features.  Zhou et al. \cite{zhuo2017indoor} introduced a method to jointly learn instance segmentation, semantic labeling and support relationships by exploiting hierarchical segmentation using a Markov Random Field for indoor RGB-D images. The inference in the MRF model is performed using an integer linear program that can be efficiently solved. 


Note that some of the approaches discussed previously under the individual sub-tasks also perform holistic reasoning. For example, \cite{silberman2012indoor} jointly models the semantic labels and physical relationships between objects, \cite{song2016semantic} jointly reconstructs 3D scene and provides voxel labels, \cite{khan2015separating} concurrently performs segmentation and cuboid detection while \cite{tejani2014latent} detects the objects and their 3D pose in a unified framework.  Such a task integration helps in incorporating wider context and results performance improvements across the tasks, however the model complexity significantly increases and efficient learning and inference algorithms are therefore required for a feasible solution.

\section{Evaluation and Discussion}
\subsection{Evaluation Metrics}

\subsubsection{Metric for classification}
Classification is the task of categorizing a scene or an object in to its relevant class. Classifier performance can be measured by classification accuracy as follows:
\begin{equation}
\textrm{Accuracy} = \frac{\textrm{Number of samples correctly classified}}{\textrm{Total number of samples}}.
\label{Classfication-formula}
\end{equation}

\subsubsection{Metric for object detection} 
Object detection is the task of recognizing each object instance and its category. Average precision (AP) is a commonly used metric to measure an object detector's performance:
\begin{equation}
\text{AP} = \frac{1}{\text{classes}}\sum _{i\in \text{classes}} \frac{TP(i)}{TP(i) + FP(i)},
\label{Object-formula}
\end{equation}
where
\begin{itemize}
\item $TP(i)$ represent the number of true positives i.e.,  predictions for class $i$ that match with the ground-truth.
\item $FP(i)$ represent the number of false positives i.e., predictions for class $i$ that do not match with the ground-truth.
\end{itemize}

\subsubsection{Metric for pose estimation} 
Object pose estimation task deals with finding object's position and orientation with respect to a certain coordinate system. The percentage of correctly predicted poses is the efficiency measure of a pose estimator. A pose estimation is consider correct if average distance between the estimated pose and ground truth is less than a specific threshold (e.g., 10\% of the object diameter).

\subsubsection{Saliency prediction evaluation metric}
Saliency prediction deals with the detection of important objects and events in a scene. There are many evaluation metrics for saliency prediction including Similarity, Normalized Scanpath Saliency ($NSS$) and F-measure ($F_{\beta}$).
\begin{equation}
\bullet\;\; \text{Similarity} = {SM} - {FM},
\label{Pose-similarity-formula}
\end{equation}
where $SM$ is the predicted saliency map and $FM$ is the human eye fixation map or ground truth.

\begin{equation}
\bullet\;\; NSS(p) = \frac{SM(p) - \mu_{SM}}{\delta_{SM}},
\label{Pose-NSS-formula}
\end{equation}
where SM is the predicted saliency map, p is the location of one fixation, $\mu_{SM}$ is the mean value of predicted saliency map and $\delta_{SM}$ is the standard deviation of the predicted saliency map.
Final $NSS$ score is the average of all $NSS(p)$ for all fixations.

\begin{equation}
\bullet\;\; F_{\beta} = \frac{\left ( 1 + \beta ^{2} \right ) * \text(percision) * \text(recall)}{\beta^{2} * \text(percision) + \text(recall)},
\label{Pose-F-formula}
\end{equation}
where $\beta ^{2}$ is a hyper parameter normally set to 0.3.

\subsubsection{Segmentation evaluation metrics and results}
Semantic segmentation is the task that involves labeling each pixel in a given image by its corresponding class. Following evaluation metrics are used to evaluate a model's accuracy:
\begin{equation}
\begin{aligned}
& \text{Pixel Accuracy}  =  \frac{\sum _{i} n_{ii}}{\sum _{i}t_{i}} \\
& \text{Mean Accuracy} = \left ( \frac{1}{^{n_{cl}}} \right )\sum _{i}\frac{n_{ii}}{t_{i}} \\
& \textrm{MIoU} = \left ( \frac{1}{^{n_{cl}}} \right )\sum _{i}\frac{n_{ii}}{\left ( t_{i} + \sum _{j} n_{ji} - n_{ii}\right )} \\
& \textrm{FIoU} = \left ( \sum _{k} t_{k}\right )^{-1}  \sum _{i} \frac{t_{i} n_{ii}}{\left ( t_{i} + \sum _{j}n_{ji} - n_{ii} \right )},
\end{aligned}
\end{equation}
where, MIoU stands for Mean Intersection over Union, FIoU denotes Frequency weighted Intersection over Union, $n_{cl}$ are the number of different classes, $n_{ii}$ is the number of pixels of class $i$ predicted to belong to class $i$, $n_{ji}$ is the number of pixels of class $i$ predicted to belong to class $j$ and $t_{i} = \sum_{j} n_{ji}$ is the total number of pixels belong to class $i$.

\subsubsection{Affordance prediction evaluation metric}
Affordance is the ability of a robot to predict possible actions that can be performed on or with an object. The common evaluation metric for affordance is the accuracy:
\begin{equation}
\textrm{Accuracy} = \frac{\textrm{Number of affordances correctly classified}}{\textrm{Total number of affordances}}.
\label{Affordance-formula}
\end{equation}

\subsubsection{3D Reconstruction}
3D reconstruction is a task of recovering full 3D shape from a single or multiple RGB-D images. Intersection over union is commonly used as an evaluation metric for the 3D reconstruction task,
\begin{equation}
\textrm{IoU} = \sum _{i}\frac{v_{ii}}{\left ( T_{i} + \sum _{j} v_{ji} - v_{ii}\right )} .
\end{equation}
where $v_{ii}$ is the number of voxels of class $i$ predicted to belong to class $i$, $v_{ji}$ is the number of voxels of class $i$ predicted to belong to class $j$ and $T_{i} = \sum_{j} v_{ji}$ is the total number of voxels belong to class $i$.

\begin{table}[ht]
\centering
\caption{Performance comparison between prominent 3D object classification methods on ModelNet40 dataset \cite{wu20153d}.}
\label{Classification-table}
\setlength\tabcolsep{1.0pt}
\begin{tabular}{@{}L{2.0cm} L{5.8cm} c@{}}
\toprule
Method &  Key Point  & Accuracy \\ \midrule

Wu et al. \cite{wu20153d} & Feature learning from 3D voxelized input  & 77.3  \\ 
Wu et al. \cite{wu2016learning}& Unsupervised feature learning using 3D generative adversarial modeling & 83.3 \\ 
Qi et al. \cite{qi2016pointnet} & Point-cloud based representation learning &  86.2  \\ 
Su et al. \cite{su2015multi} & Multi-view 2D images for feature learning & 90.1  \\  
Qi et al. \cite{qi2016volumetric} & Volumetric and multi-view feature learning &  91.4  \\ 
Brock et al. \cite{brock2016generative} & Generative and discriminative voxel modeling & 95.5 \\ \bottomrule

\end{tabular}
\end{table}

\begin{table}[ht]
\centering
\caption{Performance comparison between state-of-the-art 3D object detection methods}
\label{Object-Detection-table}
\setlength\tabcolsep{1.0pt}
\centering
\begin{tabular}{@{}L{2cm} L{5.0cm} p{1.0cm} c@{}}
\toprule
 Method  &  Key Point & Dataset  & mAP \\ \midrule
Song et al. \cite{song2016deep} & 3D CNN for amodal object detection & \multirow{4}{*}{\parbox[t]{1.0cm}{SUN RGB-D \cite{song2015sun}}} & 42.1 \\ 
 Lahoud and Ghanem \cite{lahoud20172d}   & Using 2D detection algorithms for 3D object detection & & 45.1 \\ 
 Ren et al. \cite{ren2016three} & Based on oriented gradient descriptor &  & 47.6 \\
 Qi et al. \cite{qi2017frustum} & Processing raw point cloud using CNN &  & 54.0 \\ 
 \midrule
  Song et al. \cite{song2016deep} & 3D CNN for amodal object detection & \multirow{2}{*}{\parbox[t]{1.0cm}{NYUv2 \cite{NYUDv2}}} & 36. 3 \\ 
 Deng et al. \cite{deng2017amodal} & Two-stream CNN with 2D object proposals for 3D amodal  detection. &  & 40.9 \\ \bottomrule
\end{tabular}
\end{table}













\begin{table*}[ht]
\centering
\caption{Performance of state-of-art segmentation methods on NYU v2 \cite{NYUDv2} dataset}
\label{segmentation-table}
\begin{tabular}{@{} L{3.2cm} L{7.5cm} L{1.2cm} L{1.2cm} L{1.5cm}c@{}}
\toprule
 Method  &  Key Point & No. of classes & Pixel Accuracy & Mean Class Accuracy & Mean IOU \\ \midrule

Silberman et al. \cite{NYUDv2} & Hand-crafted features (SIFT) & 4 & 58.6 & - &  - \\ 

Couprie et al. \cite{couprie2014toward} & CNN as feature extractor used with super-pixels &  4 & 64.5 & 63.5 & - \\ 

Couprie et al. \cite{couprie2014toward} & CNN as feature extractor used with super-pixels  & 13 & 52.4 & 36.2 & - \\ 

Hermans et al. \cite{hermans2014dense} & 2D to 3D label transfer and use of 3D CRF & 13 & 54.2 & 48.0 & -\\

Wolf et al. [193] & 3D decision forests & 13 & 64.9 & 55.6 & 39.5 \\ 

Tchapmi et al. \cite{tchapmi2017segcloud} & CNN with fully connected CRF & 13 & 66.8 & 56.4 & 43.5 \\ 

Gupta et al. \cite{Gupta2014} & CNN as feature extractor with novel depth embedding & 40 & 60.3 & - & 28.6 \\

Long et al. \cite{long2015fully} & Encoder decoder architecture with FCN & 40 & 65.4 & 46.1 & 34.0 \\ 

Badrinarayanan et al. \cite{badrinarayanan2015segnet} & Encoder decoder architecture with reduced parameters & 40 & 66.1 & 36.0 & 23.6 \\ 

Kendall et al. \cite{kendall2015bayesian}& Encoder decoder architecture with probability estimates & 40 & 68.0 & 45.8 & 32.4 \\ \bottomrule

\end{tabular}
\end{table*}

\begin{table*}[ht]
\centering
\caption{Performance of state-of-the-art methods for 3D reconstruction through scene completion on Rendered NYU \cite{firman2016structured}. }
\label{scene-completion-table}
\setlength\tabcolsep{2.0pt}
\begin{tabular}{@{} L{2.5cm} L{8.5cm} L{3cm} L{1.2cm} L{1.2cm} c@{}}
\toprule
Method   & Key Point & Train set & Precision & Recall & IoU \\ \midrule
Zheng et al. \cite{zheng2013beyond} & Based on geometric and physical reasoning  & \multirow{3}{*}{\parbox[t]{2.0cm}{Rendered NYU \cite{firman2016structured}}} & 60.1 & 46.7 & 34.6 \\ 
Firman et al. \cite{firman2016structured} & Estimating occluded voxels in a scene by comparison with a similar scene  &  & 66.5 & 69.7 & 50.8 \\

Song et al. \cite{song2016semantic} &  Dilation CNN with joint semantic labeling for scene completion &  & 75.0 & 92.3 & 70.3 \\ 

Song et al. \cite{song2016semantic} & Dilation CNN with joint semantic labeling for scene completion  & Rendered NYU \cite{firman2016structured} + SUNCG \cite{song2016ssc} & 75.0 & 96.0 & 73.0 \\ \bottomrule
\end{tabular}
\end{table*}



\begin{table}[ht]
\centering
\caption{Performance of state-of-art saliency detection methods on LSFD \cite{li2014saliency} dataset}
\label{Saliency-detection-table}
\setlength\tabcolsep{1.0pt}
\begin{tabular}{@{}L{2.2cm} L{5.3cm} c@{}}
\toprule
Method  &  Key Point & F-measure\\ \midrule
He et al. \cite{he2015supercnn}  & Super-pixels with CNN  &   0.698 \\ 
Zhang et al.  \cite{zhang2015minimum}  & Using min. barrier distance transform   &  0.730  \\ 
Qin et al.  \cite{qin2015saliency}  & Dynamic evolution modeling   &    0.731 \\ 
Wang et al. \cite{wang2015deep}& Based on local and global features  & 0.738\\
Peng et al. [168]& Fusion of depth modality with RGB    & 0.704\\
Ju et al. \cite{ju2014depth}& Based on  anisotropic center-surround diff. &  0.757\\
Ren et al. \cite{ren2015exploiting} & Based on depth and normal priors & 0.788\\
Feng et al. [175]& Local background enclosure based feature & 0.712\\
Qu et al. [176] & Fusing engineered features with CNN &  0.844\\ \bottomrule

\end{tabular}
\end{table}

\subsection{Discussion on Results}
In this section, we present quantitative comparisons on a set of key sub-tasks including shape classification, object detection, segmentation, saliency prediction and 3D reconstruction. Wu et al. \cite{wu20153d} created a 3D dataset, ModelNet40, for shape classification which is publicly available. Since then a number of algorithms have been proposed and tested on this dataset. Performance comparison is shown in Table~\ref{Classification-table}. It can be seen that most successful methods use CNN to extract features from 3D voxelized or 2D multi-view representations and the method based on generative and discriminative modeling \cite{brock2016generative} outperformed other competitors on this dataset. Object detection results are shown in Table~\ref{Object-Detection-table}. Point cloud is normally a difficult to process data representation and therefore, it is usually converted to other representations such as voxel or octree before further processing. However, an interesting result from Table~\ref{Object-Detection-table} is that processing point clouds directly using CNNs can boost the performance. Next comparison is shown in Table~\ref{segmentation-table} for semantic segmentation task on NYU \cite{NYUDv2} dataset. It is evident that encoder-decoder network architecture with a measure of uncertainty \cite{kendall2015bayesian} outperforms for RGBD semantic segmentation task. Saliency prediction algorithms are compared against LSFD \cite{li2014saliency} in Table \ref{Saliency-detection-table} where the most promising results are delivered when hand-crafted features are combined with CNN feature representation \cite{qu2017rgbd}. 3D reconstruction algorithms are compared in Table \ref{scene-completion-table} where again best performing method is based on CNN that effectively incorporates contextual information using dilated convolutions. As a general trend, we note that context plays a key role in individual scene understanding tasks as well as holistic scene understanding. Several approaches to incorporate scene context have been proposed e.g., skip connections in encoder-decoder frameworks, dilated convolutions, combination of global and local features. Still, the encoding of useful scene context is an open research problem.

\section{Challenges and Future Directions}
\textbf{Light-weight Models:} In the last few years, we have seen a dramatic growth in the capabilities of computational resources for machine vision applications \cite{NvidiaGPU}.  However, the deployment of large-scale models on hand-held devices still faces several challenges such as limited processing capability, low memory and power resources. The design of practical systems require a careful consideration about model complexity and desired performance. It also demands development of novel light-weight deep learning models, highly parallelize-able algorithms and compact representations for 3D data.

\textbf{Transfer Learning:} Scene understanding involves prediction about several inter-related tasks, e.g., semantic labeling can benefit from scene categorization and object detection and vice versa. The basic tasks such as scene classification, scene parsing and object detection have normally large quantities of annotated examples available for training, however, other tasks such as affordance prediction, support prediction and saliency prediction do not have huge datasets available. A natural choice for these problems is to use the knowledge acquired from pre-training performed on  a large-scale 2D, 2.5D or 3D annotated dataset. However, the two domains are not always closely related and it is an open problem to optimally adapt an existing model to the desired task such that the knowledge is adequately transfered to the new domain.

\textbf{Emergence of Hybrid Models:} Holistic scene understanding requires high flexibility in the learned model to incorporate domain knowledge and priors based on previous history of experiences and interactions with the physical world.  Furthermore, it is often  required to model wide contextual relationships between super-pixels, objects, labels or among similar type scenes to be able to reason about more complex tasks. Deep networks have turned out to be an excellent resource for automatic feature learning, however they only allow limited flexibility. We foresee a growth in the development of hybrid models, which take advantage from complementary strengths of model classes to better learn the contextual relationships. 

\textbf{Data Imbalance:} In several scene understanding tasks such as semantic labeling, some class representations are scarce while others have abundant examples. Learning a model which respects both type of categories and equally performs well on frequent as well as less frequent ones remains a challenge and needs further investigation. 

\textbf{Multi-task Learning:} Given the multi-task nature of complete scene understanding, a suitable but less investigated paradigm is to jointly train models on a number of end-tasks. As an example, for the case of semantic instance segmentation, one approach could be jointly regress the object instance bounding box, its fore-ground mask and the category label for each box. Such a formulation can allow learning robust models without undermining the performance on any single task. 

\textbf{Learning from Synthetic Data:} The availability of large-scale CAD model libraries and impressive rendering engines have provided huge quantities of synthetic data (esp. for indoor environments). Such data eliminates the extensive labeling requirements required for real data, which is a  bottleneck for training large-scale data-hungry deep learning models \cite{song2016semantic}.  Recent studies show that models trained on synthetic data can achieve strong performance on real data \cite{mccormac2016scenenet, mccormac2017scenenet}.

\textbf{Multi-modal Feature Learning:} Joint feature learning across different sensing modalities has been investigated in the context of outdoor scenes (e.g., using LIDAR and stereo cameras \cite{cadena2016multi}), but not for the case of indoor scenes. Recent sensing devices such as Matterport allow collection of multiple modalities (e.g., point clouds, mesh and depth data) in the indoor environments. Among these modalities, some have existing large-scale pre-trained models which are unavailable for other modalities. An open research problem is to leverage the frequently available data modalities and perform cross-modality knowledge transfer \cite{2017arXiv170201105A}.

\textbf{Robust and Explainable Models:} With the adaptability of deep learning models in safety critical applications including self-driving cars, visual surveillance and medical field, there comes a responsibility to evaluate and explain the decision-making process of these models. We need to develop easy to interpret frameworks to better understand decision-making of deep learning systems like in \cite{frosst2017distilling}, where Frosst et al. explained CNN decision-making using a decision tree. Furthermore, deep learning models have shown vulnerability to adversarial attacks. In these attacks, carefully-perturbed inputs are designed to mislead the model at inference time \cite{yuan2017adversarial}. There is not only a need to develop methods that can actively detect and alarm against adversarial attacks, but better adversarial training mechanisms are also required to make model robust against these vulnerabilities.



\ifCLASSOPTIONcaptionsoff
  \newpage
\fi

\bibliographystyle{IEEEtran}
\bibliography{survey}

\begin{IEEEbiography}[{\includegraphics[width=1in,height=1.25in,clip,keepaspectratio]{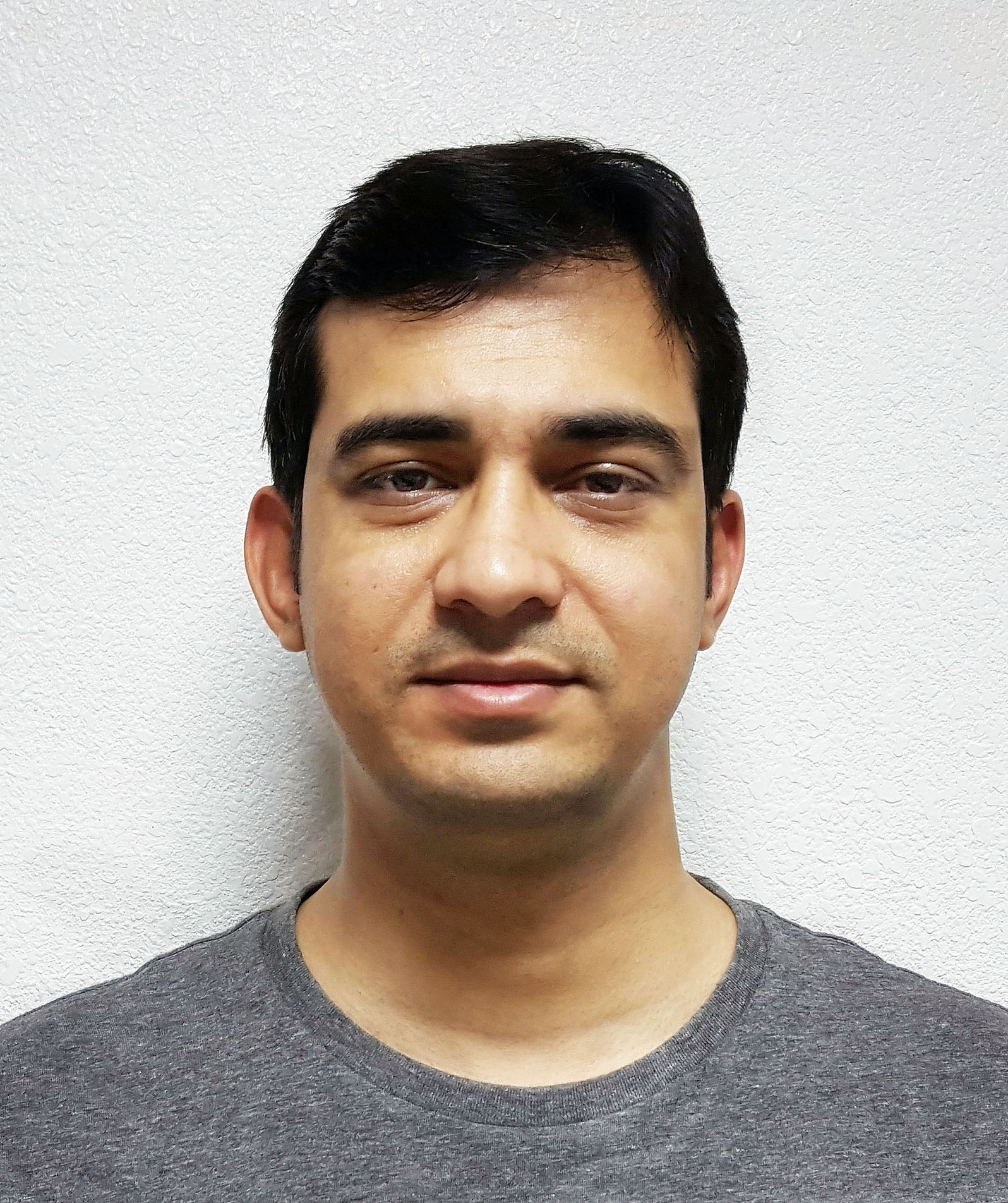}}]{Muzammal Naseer}
is a PhD candidate at Australian National University (ANU) where he is recipient of a competitive postgraduate scholarship. He received his Master degree in Electrical Engineering from King Fahd University of Petroleum and Minerals (KFUPM). He was awarded fully funded Master of science scholarship from Ministry of Higher Education, Saudi Arabia. He has also received Gold medal for outstanding performance and first class honors in BSc Electrical Engineering from the University of Punjab. His current research interests are computer vision and machine learning.
\end{IEEEbiography}

\begin{IEEEbiography}[{\includegraphics[width=1in,height=1.25in,clip,keepaspectratio]{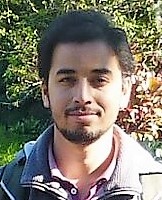}}] {Salman Khan} (M'15) received the Ph.D. degree from The University of Western Australia (UWA) in 2016. His Ph.D. thesis
received an Honorable Mention on the Dean's list
Award. He was a Visiting Researcher with National
ICT Australia, CRL, during the year 2015. He is
currently a Research Scientist with Data61 (CSIRO)
and an Adjunct Lecturer with Australian National
University (ANU) since 2016. He was a recipient of several prestigious scholarships including Fulbright and IPRS. He has served as a program committee member for several premier conferences including CVPR, ICCV and ECCV. His research interests
include computer vision, pattern recognition and machine learning.
\end{IEEEbiography}

\begin{IEEEbiography}[{\includegraphics[width=1in,height=1.25in,clip,keepaspectratio]{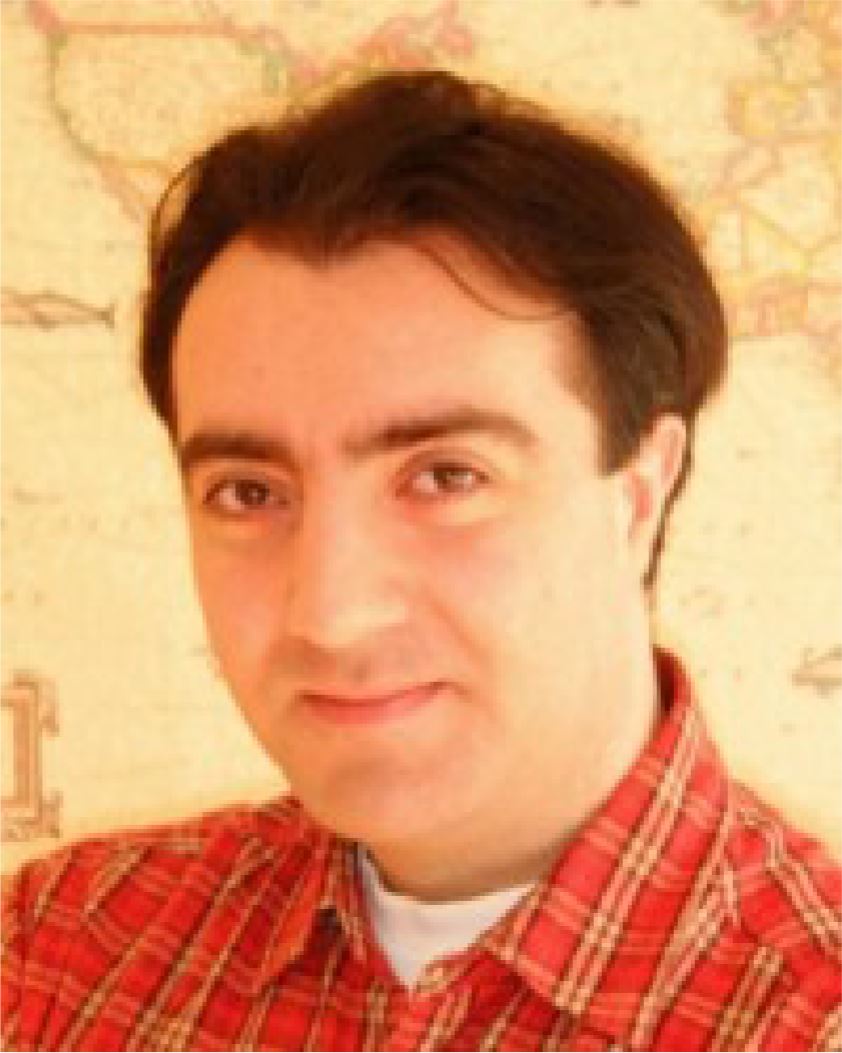}}] {Fatih Porikli} (M'96,SM'04,F'14) received the Ph.D. degree from New York University in 2002. He was the Distinguished Research Scientist with Mitsubishi Electric Research Laboratories. He is currently a Professor with the Research School of Engineering, Australian National University and a Chief Scientist
at the Global Media Technologies Lab at Huawei,
Santa Clara. He has authored over 300 publications,
coedited two books, and invented 66 patents. His
research interests include computer vision, pattern
recognition, manifold learning, image enhancement,
robust and sparse optimization and online learning with commercial applications in video surveillance, car navigation, robotics, satellite, and medical systems. He was a recipient of the Research and Development 100 Scientist of the Year Award in 2006. He received five Best Paper Awards at premier IEEE conferences and five other professional prizes. He is serving as the Associate Editor of several journals for the past 12 years. He also served at the organizing committees of several flagship conferences including ICCV, ECCV, and CVPR. He is a fellow of the IEEE.
\end{IEEEbiography}


\end{document}